\documentclass[final]{cvpr}

\usepackage{times}
\usepackage{epsfig}
\usepackage{graphicx}
\usepackage{amsmath}
\usepackage{amssymb}
\usepackage{subcaption}
\usepackage{cite}

\usepackage[pagebackref=true,breaklinks=true,colorlinks,bookmarks=false]{hyperref}

\usepackage{subfiles} 

\usepackage{chapterbib}
\usepackage{titling}

\begin{document}

\title{Spatiotemporal tomography based on scattered multiangular signals and its
application for resolving evolving clouds using moving platforms}

\author{Roi~Ronen, Yoav Y. Schechner\\
Viterbi Faculty of Electrical Engineering\\
Technion - Israel Institute of Technology\\
Haifa, Israel\\
{\tt\small roironen@campus.technion.ac.il }\\ {\tt\small yoav@ee.technion.ac.il}
\and
Eshkol Eytan\\
Department of Earth and Planetary Sciences\\
The Weizmann Institute of Science\\
Rehovot, Israel\\
{\tt\small eshkol.eytan@weizmann.ac.il}
}
\date{}
\maketitle

\begin{abstract}
    We derive computed tomography (CT) of a time-varying volumetric translucent object, using  a small number of moving cameras. 
    We particularly focus on passive scattering tomography, which is a non-linear problem. We demonstrate the approach on dynamic clouds, as clouds have a major effect on Earth’s climate. State of the art scattering CT assumes a static object. Existing 4D CT methods rely on a linear image formation model and often on significant priors. In this paper, the angular and temporal sampling rates needed for a proper recovery are discussed. If these rates are used, the paper leads to a representation of the time-varying object, which simplifies 4D CT tomography. The task is achieved using gradient-based optimization. We demonstrate this in physics-based simulations and in an experiment that had yielded real-world data.
\end{abstract}

\section{Introduction}
\label{sec:introduction}
\begin{figure}[t]
 \centering
  \includegraphics[width=1\linewidth]{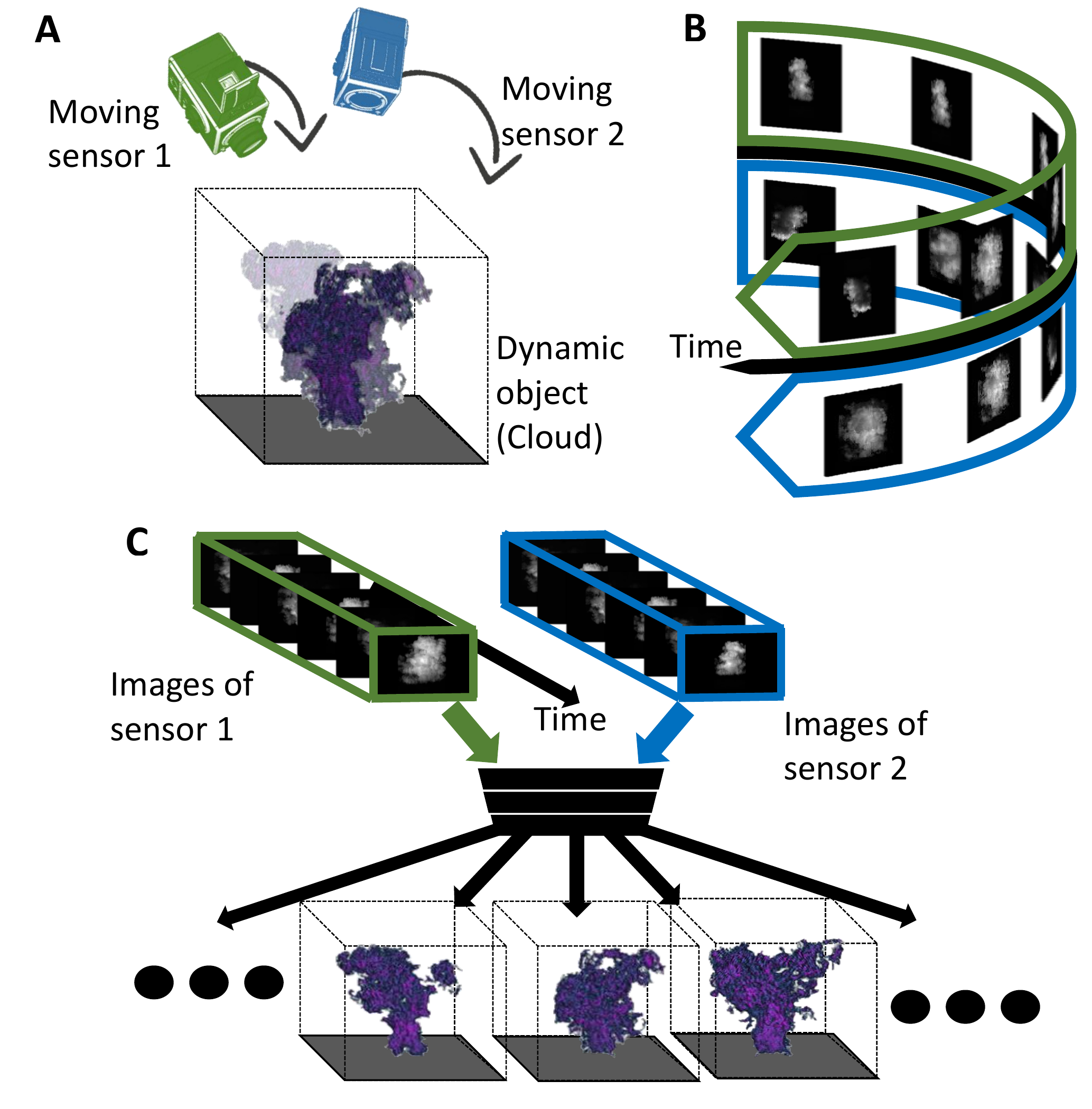}
  \vspace{-0.3cm}
  \caption{[A] Multiple moving sensors scan a time-varying object (cloud) from multiple-views. [B] The measurements,  acquired at a different times, are the input to our method, which aims to recover the 3D volume density of the object at the different times [C].}
  \label{fig:main_illustration}
\end{figure}
Computed tomography (CT) is a powerful way to recover the inner structure of three dimensional (3D) volumetric heterogeneous objects~\cite{gkioulekas2016evaluation}. Being possibly one of the earliest types of computational photography methods, CT has extensive use in many research and operational domains. These include medicine~\cite{pan20044d}, sensing of atmospheric pollution~\cite{aides2020distributed}, geophysics~\cite{wright2008scanning} and fluid dynamics~\cite{zang2019warp,zang2018space}. As a result, CT technologies and novel modalities are increasingly being advanced by the computer vision and graphics communities~\cite{gregson2012stochastic,narasimhan2005structured}. 

CT requires imaging from multiples directions~\cite{anirudh2018lose,kaestner2011spatiotemporal}. In nearly all CT approaches, the object is considered static during image acquisition. However, in many cases of interest, the object changes while  multi-view images are acquired sequentially~\cite{zang2020tomofluid,eckert2018coupled}.
Thus, effort has been invested to generalize 3D CT to four-dimensional (4D) spatiotemporal CT, particularly in the computer vision and graphics communities~~\cite{qian2017stereo,zang2020tomofluid,zang2019warp}. This effort has been directed at linear-CT modalities. Linear CT is computationally easier to handle, thus common for decades, mainly in medical imaging~\cite{hiriyannaiah1997x}. Medical CT often exploited the periodic temporal nature of organ dynamics, to synchronize sequential acquisitions~\cite{pan20044d}. The generalization in linear CT is mirrored in a generalization of object surface recovery by spatio-temporal computer vision~\cite{qian2017stereo,leroy2017multi,mustafa2016temporally}.

This paper focuses on a more complicated model: scattering CT. It is important to treat this case for scientific, societal and practical reasons.
The climate is strongly affected by interaction with clouds~\cite{fujita1986mesoscale} (Fig.~\ref{fig:main_illustration}).

To reduce major errors in climate predictions, this interaction requires a much finer understanding of cloud physics than current knowledge. Current knowledge is based on empirical remote sensing data that is analyzed under the assumption that the atmosphere and clouds are made of very broad and uniform layers. This leads to errors in climate understanding. To overcome this problem, 3D scattering CT has been suggested as a way to study clouds~\cite{levis2015airborne,levis2017multiple}. 

Scattering CT of clouds requires high resolution multi-view images from space. There are spaceborne and high-altitude systems may provide such data, such as AirMSPI~\cite{diner2013airborne}, MAIA~\cite{boland2018nasa}, HARP~\cite{neilsen2015hyper}, AirHARP~\cite{mcbride2020spatial} and the planned CloudCT formation~\cite{schilling2019cloudct}. However, there is a practical problem: these systems are very expensive, so it is not realistic to deploy them in large numbers to simultaneously acquire images of the same clouds from many angles. Therefore, in practice, the platforms are planned to move above the clouds: a sequence of images is taken, in order to span and sample a wide angular breadth, but the cloud evolves meanwhile. Hence there are important reasons to derive 4D scattering CT, particularly of clouds.  

We pose conditions under which this task can be performed. These relate to temporal sampling and angular breadth, in relation to the correlation time of the evolving object. Then, we generalize prior 3D CT, specifically scattering CT, to spatiotemporal recovery using data taken by moving cameras. We present an optimization-based method to reach this, and then demonstrate this method both in rigorous simulations and on  real data.

\section{Theoretical Background}
\label{sec:theory}

Computed Tomography (CT) seeks estimation of the 3D volumetric density of ${\ \beta}$ an object. Usually, CT  measures the object from multiple directions. Denote those measurements by ${\boldsymbol y}$. A forward model  ${\cal F} \left({\boldsymbol \beta} \right)$ expresses the image formation model. Estimation of  ${\boldsymbol \beta}$ is done by minimization of a cost ${\cal E}$, which penalizes the discrepancy between ${\boldsymbol y}$ and the forward model,
\begin{equation}
    {\hat{\boldsymbol \beta}} = \arg \!\!\min_{{\boldsymbol \beta} ~~~~~} {\cal E} \left[{\boldsymbol y}, {\cal F} \left({\boldsymbol \beta} \right) \right]
       \;.  
   \label{eq:inverse}
\end{equation}

Often, acquiring data from multiple angles simultaneously is very difficult. Sensors are expensive and/or power-consuming. Thus, their duplication in large numbers to sample many directions is often prohibitive. Sometimes the sensors are bulky and need cooling, posing difficulty to pack them densely.
Therefore, CT measurements are often acquired {\em sequentially}. In contrast, the inverse tomographic problem (Eq.~\ref{eq:inverse}) expresses the object as time-invariant. However, in many situations it is not the case;  the object changes in time. Thus, modeling  the inverse problem as Eq.~(\ref{eq:inverse}) is inconsistent both with the dynamic nature of the object and the sequential nature of the sensing process.



This paper focuses on scattering-based CT. Thus, we describe here the relevant forward model. We follow notations and definitions of~\cite{levis2015airborne,aides2020distributed}.
Material density at a voxel is denoted by $\beta$. In case the main interaction effect is scattering (rather than absorption or emission), as is the case of visible light in clouds, $\beta$ is the extinction or scattering coefficient of the medium, in units of ${\rm km}^{-1}$. Concatenating this coefficient in all voxels creates a vector ${\boldsymbol \beta}_t$, per time $t$. 
The interaction of radiation with a scattering volumetric object is modeled by 3D {\em radiative transfer}, which includes multiple scattering. Define radiative transfer by an operator ${\rm RT}({\boldsymbol \beta}_t)$. There are various algorithms to implement ${\rm RT}({\boldsymbol \beta}_t)$, including Monte-Carlo~\cite{mayer2009radiative,loeub2020monotonicity} and the spherical harmonic discrete ordinate method (SHDOM). We use the latter, as it is considered trustworthy by the scientific community~\cite{evans1998spherical} and  has open-source online codes~\cite{levis2020git}. 

Radiative transfer yields the radiance 
$i({\bf x},{\boldsymbol \omega})$ at each location ${\bf x}$ in space and each light propagation direction ${\boldsymbol \omega}$. A camera observes the scene from a specific location, while each of the camera pixels samples a direction ${\boldsymbol \omega}$.  Hence, imaging (forward model) amounts to sampling the output of radiative transfer at the camera locations and pixels' lines of slight, while integrating over the camera exposure time and spectral bands. Camera sampling is denoted by a projection operator
$P_{{\bf x},{\boldsymbol \omega}}$. To conclude, the forward model for the expected value of a pixel gray level at time $t$ is  
\begin{equation}
   g_{{\bf x},{\boldsymbol \omega},t}
   ={\cal F} \left( {\boldsymbol \beta}_{t} \right) 
     \approx \gamma^{\rm cam}
     P_{{\bf x},{\boldsymbol \omega}}
     \left\{ 
        {\rm RT}
          ({\boldsymbol  \beta}_t)
     \right\}
       \;.
   \label{eq:forwardmodel_approx}
\end{equation}
Here $\gamma^{\rm cam}$  expresses camera properties, including the lens aperture area, exposure time, spectral band, quantum efficiency and lens transmissivity. 
Eq.~(\ref{eq:forwardmodel_approx}) assumes that the exposure time is sufficiently small, such that within this time, the scene and the camera pose vary insignificantly. 

Empirical measurements include random noise. The noise mainly originates from the discrete nature of photons and electric charges, which yields a Poisson process. There are additional noise sources, and their parameters can be extracted from the sensor specifications. Denote incorporation of noise into the expected signal by the operator $\cal N$. Then, a raw measurement is 
\begin{eqnarray}
    y_{{\bf x},{\boldsymbol \omega},t} = {\cal N} 
     \left\{
       g_{{\bf x},{\boldsymbol \omega},t}
     \right\}
    \;.
  \label{eq:y_noise_I}
\end{eqnarray}

\section{Representing Dynamic Volumetric Objects}
\label{sec:3DObject}

In this section, we present an approximate representation of volumetric 3D objects, which evolve gradually in time.
There is a need and justification for the approximation. The need is because in our work, there is often insufficient  simultaneous data for high quality 3D tomography at all times. Data is captured sequentially, while the object evolves. Hence, at best, we would recover a good approximation of the evolving object. The approximation can be satisfactory, however, if temporal samples are sufficiently dense, as elaborated in Sec.~\ref{sec:FrequencyAnalysis}. 

The object has an evolving state, which is sampled at the time set ${\cal T}=\{t_1,\ t_2,\ \dots,\ t_{N^{\rm state}}\}$. At time $t\in {\cal T}$, the true object is represented by the instantaneous 3D extinction field ${\boldsymbol \beta}_t$.
Define a corresponding hidden field 
${\boldsymbol \beta}^{\rm hidden}_t$. 
The instantaneous field is represented as a convex linear combination of the hidden fields:
\begin{equation}
   {\boldsymbol \beta}_{t}= 
    \sum_{t'\in {\cal T}}
   w_{t}(t'){\boldsymbol \beta}^{\rm hidden}_{t'}
   \label{eq:beta_t_linear_comb2}
   \;.
\end{equation}
All weights $\{ w_{t}(t')\}$ satisfy
$0 \leq  w_t(t') \leq 1$
and
\begin{equation}
    \sum_{t'\in {\cal T}} w_{t}(t')=1
    \label{eq:normw}
    \;.
\end{equation}
Equation~(\ref{eq:beta_t_linear_comb2}) implies a non-negative correlation of the 3D field at $t$ to the 3D field at any time $t'$. Correlation should decay with the time lag
$|t-t'|$. 
We set the weights by a normal function
\begin{equation}
    w_t(t') = s
    \exp\left(
           -\frac{|t-t'|^2}{2\sigma^2}
        \right)
     \label{eq:weights_t_normal_dist}
      \;.
\end{equation}
Here  $s$ is a normalization factor, set to satisfy Eq.~(\ref{eq:normw}). 

The parameter $\sigma$ expresses the {\em effective correlation time} of the volumetric object. We elaborate on the value of $\sigma$ in Sec.~\ref{sec:FrequencyAnalysis}. Two limiting cases are illustrative, however. 
For $\sigma \xrightarrow{}\infty$, we have
$w_t(t')  \xrightarrow{} 1/N^{\rm state}$. This means that the 
object ${\boldsymbol \beta}$ 
is effectively static. On the other hand, for $\sigma \xrightarrow{}0$, we have
  $w_t(t')  \xrightarrow{} \delta(t-t')$, i.e. a Dirac delta function. 
This means that the 
object ${\boldsymbol \beta}$ 
varies so fast, that at any time $t$ its state is uncorrelated to the state at other times. 

\section{Bandwidth and Object Sampling}
\label{sec:FrequencyAnalysis}

In Sec.~\ref{sec:3DObject}, a 3D volumetric object which gradually evolves is represented using a linear superposition of sampled states. The linear superposition is controlled by a kernel, whose width is $\sigma$. Here we elaborate both on the sampling process and how $\sigma$ emerges from the temporal nature of the object.
The relation between a continuously varying object and its discrete temporal samples is governed by the Nyquist sampling theorem. This theorem states, that for an infinite time domain:\\
\noindent $\bullet$ Sampling loses no information if for any time sample index $k$, $|t_{k+1}-t_k|\leq (2B)^{-1}$, where $B$ is the half-bandwidth of the object's temporal variations. \\
\noindent $\bullet$ Based on these lossless temporal samples, lossless reconstruction of the object is achieved by a linear superposition of the samples. The linear superposition is achieved by temporal convolution of the samples with a ${\rm sinc}$ kernel. This kernel has infinite temporal length. The kernel's effective half-width, defined by its first zero-crossing, is $(2B)^{-1}$. \\

In practice, the temporal domain, number of samples and reconstruction kernels are finite. Moreover, the object's temporal spectrum is often not completely band-limited by 
$B$, because some low energy content has far higher frequencies. Consequently, sampling and reconstruction are lossy, yielding an approximate result. The reconstruction is not performed by a ${\rm sinc}$ kernel, but using a finite-length kernel, such as the cropped Gaussian $w_t(t')$ of Eq.~(\ref{eq:weights_t_normal_dist}). Correspondingly, in our approximation, $\sigma\sim (2B)^{-1}$.

As an example, consider warm convective clouds. They are governed by air turbulence of decameter scale. In these scales~\cite{fujita1986mesoscale}, the {\em correlation time} of content in a voxel is about $\approx 20 ~{\rm to}~50$ seconds. This indicates that {\em 4D spatiotemporal clouds can be recovered well using 4D spatiotemporal samples}, if
the temporal samples are about 30 seconds apart. Furthermore,
this indicates the range of values of $\sigma$. Moreover, the entire lifetime of a warm convective cloud is typically measured in minutes.

\begin{figure}[t]
 \centering
  \includegraphics[width=1\linewidth]{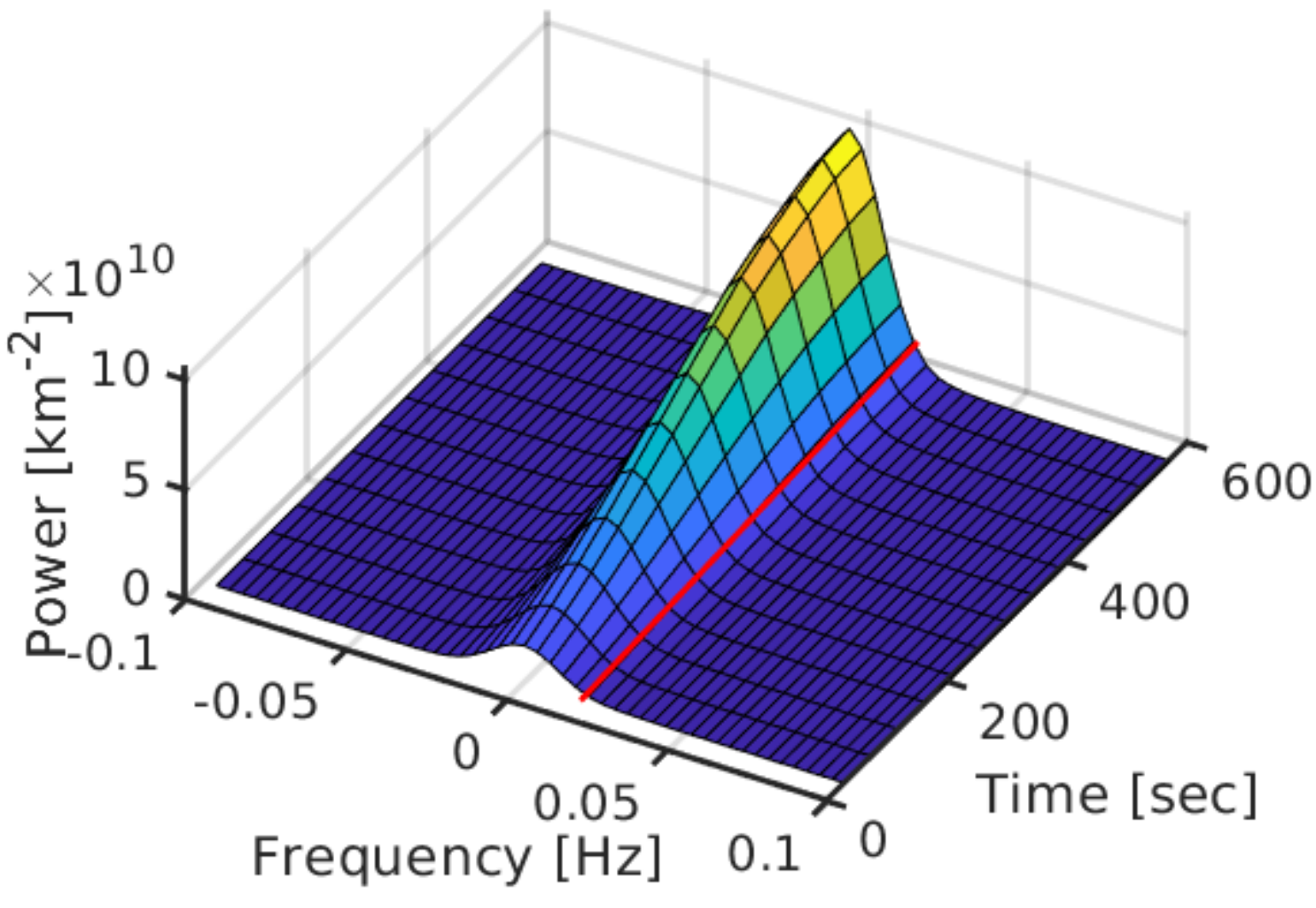}
  \vspace{-0.7cm}
  \caption{A cutoff frequency $\approx 1/50 {\rm [Hz]}$, within which 95\% of the signal power is contained, is marked in red.}
  \label{fig:stft}
\end{figure}

An additional illustration is given by a cloud simulation, which is described in detail in Sec.~\ref{sec:CloudFieldSimulation}. The cloud evolves for about 10 minutes. For each cloud voxel, we calculated the power spectrum using the short-time Fourier transform. This power was then aggregated over all voxels. The total temporal spectrum is plotted in Fig~\ref{fig:stft}.  
The spectrum is effectively limited. The cutoff is not sensitive to evolving stages of the clouds. From this cutoff, a temporal sampling period which is $25{\rm [sec]}$ or shorter encapsulates most of the energy of the temporal variations. Hence, we set $\sigma \approx 25{\rm [sec]}$ for clouds. 

\section{Tomographic Angular Extent}
\label{sec:anglesmaple}

Section~\ref{sec:FrequencyAnalysis} dealt with sampling of an object, as if 4D measurements are done in-situ. However, in CT, we have no direct access to ${\boldsymbol \beta}_t$: we only measure  projections ${\boldsymbol y}_{t}$. As we discuss now, projections must have a {\em wide angular breadth, while object evolution is small}. 

\begin{figure}[t]
    \centering
        \includegraphics[width=0.8\linewidth]{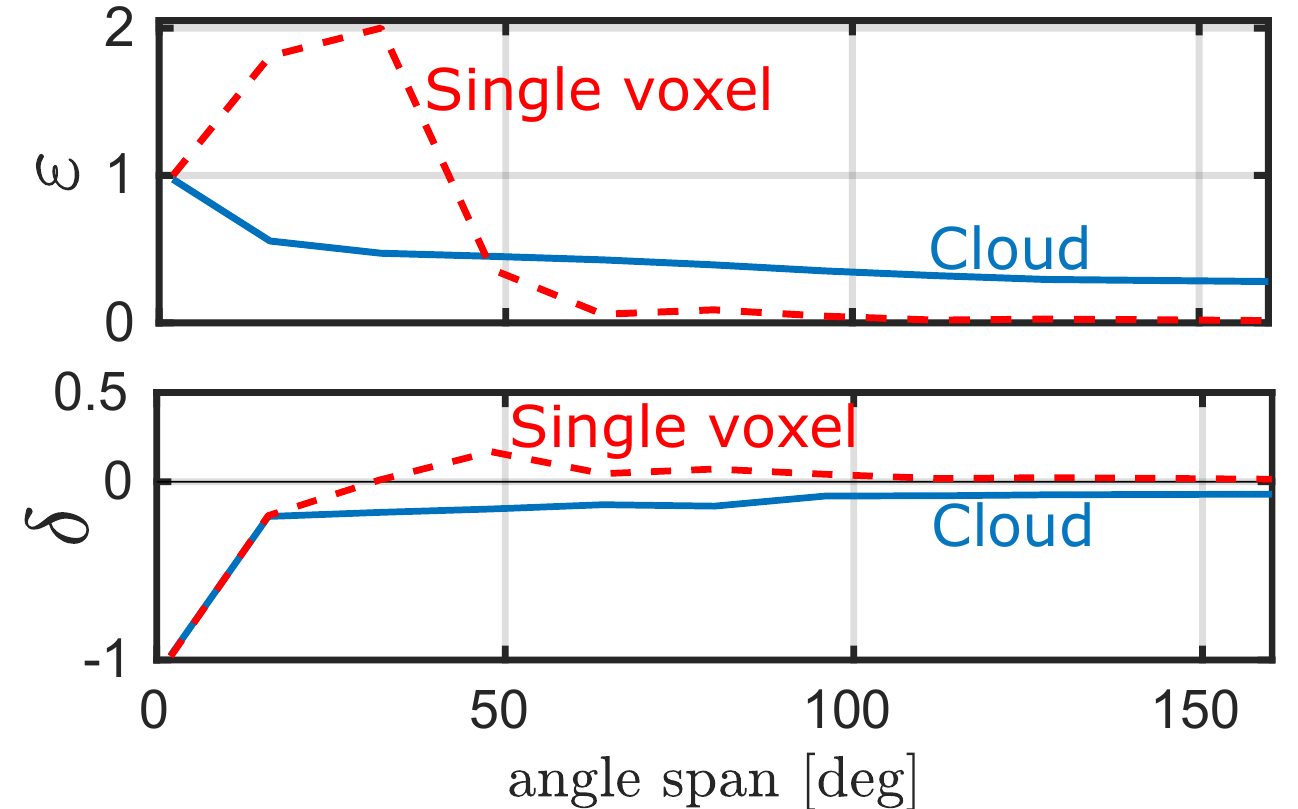}
        \vspace{-0.3cm}
    \caption{A static heterogeneous cloud and a single-voxel cloud (having size $20 {\rm m}\times 20 {\rm m}\times 20 {\rm m}$) are recovered from nine viewpoints. The plots are of errors defined in Eq.~(\ref{eq:epsilin}).}
    \label{fig:angle_span}
  \end{figure}

Consider an extreme case. Let a cloud be temporally constant and reside only in a single voxel, over the ocean. Viewed from space by two cameras simultaneously, cloud recovery here amounts to triangulation. In triangulation, the best cloud-localization resolution is obtained if the angular range between the two cameras is $90^{\circ}$. At small baselines, localization decreases linearly with the decreasing angular extent. 
When more than two cameras operate, localization behaves more moderately, but with a similar trend. Consider two criteria that had been used in 3D CT~\cite{levis2015airborne,loeub2020monotonicity,holodovsky2016situ}. Per time sample $t$, 
\begin{equation}
    \delta_t =
       \frac{\| {\boldsymbol \beta}^{\rm true}_t\|_1 - \|{\hat {\boldsymbol \beta}}_t \|_1}{\| {\boldsymbol \beta}^{\rm true}_t\|_1}
       \;,\;\;\;
       \varepsilon_t = \frac{\| {\boldsymbol \beta}^{\rm true}_t - {\hat {\boldsymbol \beta}}_t \|_1}{\| {\boldsymbol \beta}^{\rm true}_t\|_1}
       \;
\label{eq:MassError}
\end{equation}
relate, respectively, to the relative bias of the object mass and the relative recovery error.  We generalize them to the whole sample set 
$t\in{\cal T}$, by averaging:
\begin{equation}
   \delta = 
   \frac{1}{N^{\rm state}}
   \sum_{t\in{\cal T}}  \delta_t 
   \;, \;\;\;\;
   \varepsilon = 
   \frac{1}{N^{\rm state}}
   \sum_{t\in{\cal T}}  \varepsilon_t 
   \;.
   \label{eq:epsilin}
\end{equation}
Fig.~\ref{fig:angle_span} plots the measures when CT attempts to recover a single-voxel cloud (extending 20 meters), when 9 cameras surround it from 500km away. 
Above $\approx 60^{\circ}$ total angular extent, recovery reaches a limiting excellent quality, but quality is very poor at narrow angle spans. 

What happens in extended objects? In linear-CT (as in medical X-ray CT), information loss due to limited-angle imaging is known as the {\em missing cone of frequencies}~\cite{macias1988missing,agard1989fluorescence}. In scattering CT, with the exception of very sparse objects, the missing cone linear theory does not apply. However (See Fig.~\ref{fig:angle_span}), there is a marked degradation of quality if the angular extent is narrow. Here, scattering CT is performed on a single-state (fixed time) of a cloud simulated as in Sec.~\ref{sec:CloudFieldSimulation}.

So far, this section dealt with static clouds. Clouds are considered nearly static between times $t,t'$ if $|t-t'|<\sigma$. 
The viewing angular extent covered in those times (and in intermediate times) is denoted $\Theta(t,t')$, in radians. So, within time span $\approx \sigma$ good recovery can be achieved only if $2\Theta(t,t')/\pi$ is large. If it is low, then spatial (altitude) resolution in CT recovery is lost.  Most CT systems cover wide angular extent, eventually. So, guarantee for quality is the angular rate. Define a dimensionless figure
\begin{equation}
   \rho =\frac{2\Theta(t,t')}
              {\pi}
          \frac{\sigma}
              {|t-t'|}
   \;.
   \label{eq:rho}
\end{equation}
Good 4D recovery requires $\rho\gtrsim 1$, while temporal sampling satisfies the condition of Sec.~\ref{sec:anglesmaple}. The more these conditions are violated, the worse 4D CT is expected to perform.


\section{4D Scattering Tomography}
\label{sec:4DTomography}

We now generalize Eq.~(\ref{eq:inverse}) to 4D CT.  At any time $t\in{\cal T}$, the object is viewed simultaneously from a set of viewpoints ${\cal C}_t$. The image data captured in viewpoint $c\in{\cal C}_t$ is denoted by the vector ${\boldsymbol y}_{c}$. The image data captured simultaneously by all cameras  $c\in{\cal C}_t$ is concatenated to a vector ${\boldsymbol y}_{t}$. At that time, the modeled medium variables are represented by a vector 
${\boldsymbol \beta}_t$. At the corresponding time, as described in Sec.~\ref{sec:3DObject}, 
there is a hidden representation of the medium, 
${\boldsymbol \beta}^{\rm hidden}_t$.

The inverse problem is now formulated by
\begin{equation}
     \{   
        {\hat {\boldsymbol \beta}}_t 
     \}_{t\in {\cal T}}
     = 
      \arg \!\!\!\!\!\!\!\!\min_{
       \{   
           {\boldsymbol \beta}_t 
        \}_{t\in {\cal T}}
       ~~~~~
       } 
       \sum_{t\in {\cal T}}
       {\cal E} 
       \left[{\boldsymbol y}_{t}, {\cal F} \left({\boldsymbol \beta}_t \right) \right]
       \;.  
   \label{eq:dynamic_inverse}
\end{equation}
We use
\begin{equation}
       {\cal E} 
       \left[{\boldsymbol y}_{t}, {\cal F} \left({\boldsymbol \beta}_t \right) \right]
       =
       \frac{1}{2}
       \|
        {\boldsymbol y}_{t} - {\cal F} \left({\boldsymbol \beta}_t \right)     
       \|^2_2
       \;.  
   \label{eq:Et}
\end{equation}
Eq.~(\ref{eq:dynamic_inverse}) can be solved efficiently by gradient-based methods.
Towards this, let us approximate the gradient of the cost being minimized in Eq.~(\ref{eq:dynamic_inverse}) by
\begin{equation}
      \frac{\partial }
           {\partial {\boldsymbol \beta}_t}
       \sum_{t'\in {\cal T}}
       {\cal E} 
       \left[{\boldsymbol y}_{t'}, {\cal F} \left({\boldsymbol \beta}_{t'} \right) \right]
       \approx 
       \frac{\partial }
           {\partial {\boldsymbol \beta}^{\rm hidden}_t}
       \sum_{t'\in {\cal T}}
       {\cal E} 
       \left[{\boldsymbol y}_{t'}, {\cal F} \left({\boldsymbol \beta}_{t'} \right) \right]
       \;.  
   \label{eq:smigrad}
\end{equation}
Note that
\begin{eqnarray}
      \frac{\partial }
           {\partial {\boldsymbol \beta}^{\rm hidden}_t}
       \sum_{t'\in {\cal T}}
       {\cal E} 
       \left[{\boldsymbol y}_{t'}, {\cal F} \left({\boldsymbol \beta}_{t'} \right) \right]
       = ~~~~~~~~~~~~~~~~~~~~~
       \nonumber \\ ~~
       \sum_{t'\in {\cal T}}
     \frac{\partial 
          {\cal E} 
          \left[{\boldsymbol y}_{t'}, {\cal F} 
             \left({\boldsymbol \beta}_{t'} \right) 
          \right]
          }
          {\partial {\boldsymbol \beta}_{t'}}
     \frac{\partial {\boldsymbol \beta}_{t'}}
          {\partial {\boldsymbol \beta}^{\rm hidden}_t}
       \;.  
   \label{eq:grandgrad}
\end{eqnarray}
From Eq.~(\ref{eq:Et}),
\begin{equation}
   \frac{\partial 
        {\cal E} 
          \left[{\boldsymbol y}_{t'}, {\cal F} 
            \left({\boldsymbol \beta}_{t'} \right) 
          \right]
       }
       {\partial {\boldsymbol \beta}_{t'}}
    =
   \left[ 
      {\cal F} \left({\boldsymbol \beta}_{t'} \right)  - {\boldsymbol y}_{t'} 
   \right]      
   \frac{\partial 
         {\cal F} \left({\boldsymbol \beta}_{t'} \right)
        }
        {\partial {\boldsymbol \beta}_{t'}},
   \label{eq:derivative_eps_t}
\end{equation}
while from Eq.~(\ref{eq:beta_t_linear_comb2}), 
\begin{equation}
    \frac{\partial {\boldsymbol \beta}_{t'}}
          {\partial {\boldsymbol \beta}^{\rm hidden}_t}
    =w_{t'}(t) \;.          
    \label{eq:dbetadbeta}
\end{equation}
Define the set of medium density fields at all sampled times 
 ${\cal B}= \{ {\boldsymbol \beta}_{t'}\}_{t'\in {\cal T}} $.
From Eqs.~(\ref{eq:smigrad},\ref{eq:grandgrad},\ref{eq:derivative_eps_t},\ref{eq:dbetadbeta}),  for optimizing the field at time $t$, the approximate gradient is
\begin{equation}
      {\bf g}_t({\cal B}) = \sum_{t'\in {\cal T}}
      w_{t'}(t)
     \left[ 
        {\cal F} \left({\boldsymbol \beta}_{t'} \right)  - {\boldsymbol y}_{t'} 
     \right]      
     \frac{\partial 
          {\cal F} \left({\boldsymbol \beta}_{t'} \right)
          }
          {\partial {\boldsymbol \beta}_{t'}}
       \;.  
   \label{eq:pgrad}
\end{equation}
A gradient-based approach then performs per iteration $k$:
\begin{equation}
  {\boldsymbol \beta}_{t}(k+1)=
  {\boldsymbol \beta}_{t}(k)-\alpha {\bf g}_t({\cal B}_k)
  \label{eq:generic_gradient_descent}
\end{equation}
where $\alpha$ is a step size and 
 ${\cal B}_k= 
   \{ 
      {\boldsymbol \beta}_{t'}(k)
   \}_{t'\in {\cal T}} $.

The approach in Eqs.~(\ref{eq:dynamic_inverse}-\ref{eq:generic_gradient_descent}) is not specific to scattering CT. The formulation {\em can apply generically to other inverse problems} where data is acquired sequentially while the object evolves, and the forward model ${\cal F}$ 
is known and differentiable. In case of scattering CT,  ${\cal F}$  is discussed in Sec.~\ref{sec:theory}. Computing the Jacobian 
$     \partial {\cal F} \left({\boldsymbol \beta}_{t'} \right)
        /
         \partial {\boldsymbol \beta}_{t'}$
is complex then. However, there are approximations to the Jacobian of 3D RT, which  can be computed efficiently~\cite{levis2015airborne,loeub2020monotonicity}, making the recovery tractable. 
The complexity of solving Eq.~(\ref{eq:dynamic_inverse}) is similar 
to 3D static CT by (\ref{eq:inverse}). We performed optimization using a L-BFGS-B solver~\cite{zhu1997algorithm}. 

Following~\cite{levis2017multiple}, prior to the optimization process, the set of voxels to estimate is bounded using space-carving~\cite{kutulakos2000theory}. Space-carving bounds a 3D shape by back-projecting multi-view images. A voxel is labeled as belonging to the object, if the number of back-projected rays that intersect this voxel is greater than a threshold. To adapt this approach to our dynamic framework, the shape was estimated in a coarse spatial grid and using a low threshold for labeling voxels as potentially part of the cloud.

\section{Simulations} 
\label{sec:Simulations}


We test the proposed method on clouds. The atmosphere is a scattering medium which changes continuously. It  includes several types of scattering particles, including water droplets, aerosols and molecules. Scattering by water droplets is usually much more dominant and spatiotemporally variable than aerosols, hence we focus on the former. Scattering by air molecules does not require imaging: it follows  Rayleigh theory. Molecular density changes mainly in height and is usually known globally using non-imaging sensors. Thus, the evolving concentration of cloud water droplets is the main unknown we sense and seek. 

\subsection{Cloud Simulation} 
\label{sec:CloudFieldSimulation}

For realistic complexity, we use a rigorous simulation based on cloud physics, including evolution of the cloud microphysics. Clouds are simulated using the
System of Atmospheric Modeling (SAM)~\cite{khairoutdinov2003cloud}, which is a non-hydrostatic, inelastic large eddy simulator (LES)~\cite{neggers2003size,xue2006large,heus2009statistical}. It describes the turbulent atmosphere using equations of momentum, temperature, water mass balance and continuity. This model was coupled to a spectral (bin) microphysical model (HUJI SBM)~\cite{khain2004simulation,fan2009ice} of the droplets' size. It propagates the evolution of the droplet's size distribution by solving the equations for nucleation, diffusional growth, collision-coalescence and break-up. This is done on a logarithmic grid of 33 bins $[2 \mu {\rm m},3.2 {\rm mm}]$. 

The simulation runs according to the BOMEX case~\cite{siebesma2003large} of trade wind cumulus clouds near Barbados. Humidity and potential temperature profiles are used as initial conditions, while the surface fluxes and large-scale forcing are constant. The mean horizontal background wind is zero.
The horizontal boundary condition is cyclic. The domain is 5.12km long (cloud diameter is $\approx 800{\rm m}$) at 10m resolution. Vertically, the resolution is 10m from sea level to 3km, and then resolution coarsens to 50m. The cloud top reaches 2km. The simulation expresses an hour, of which 30 minutes includes the cloud's lifetime. The temporal resolution is 0.5sec.  

\begin{figure}[t]
 \centering
  \includegraphics[width=1\linewidth]{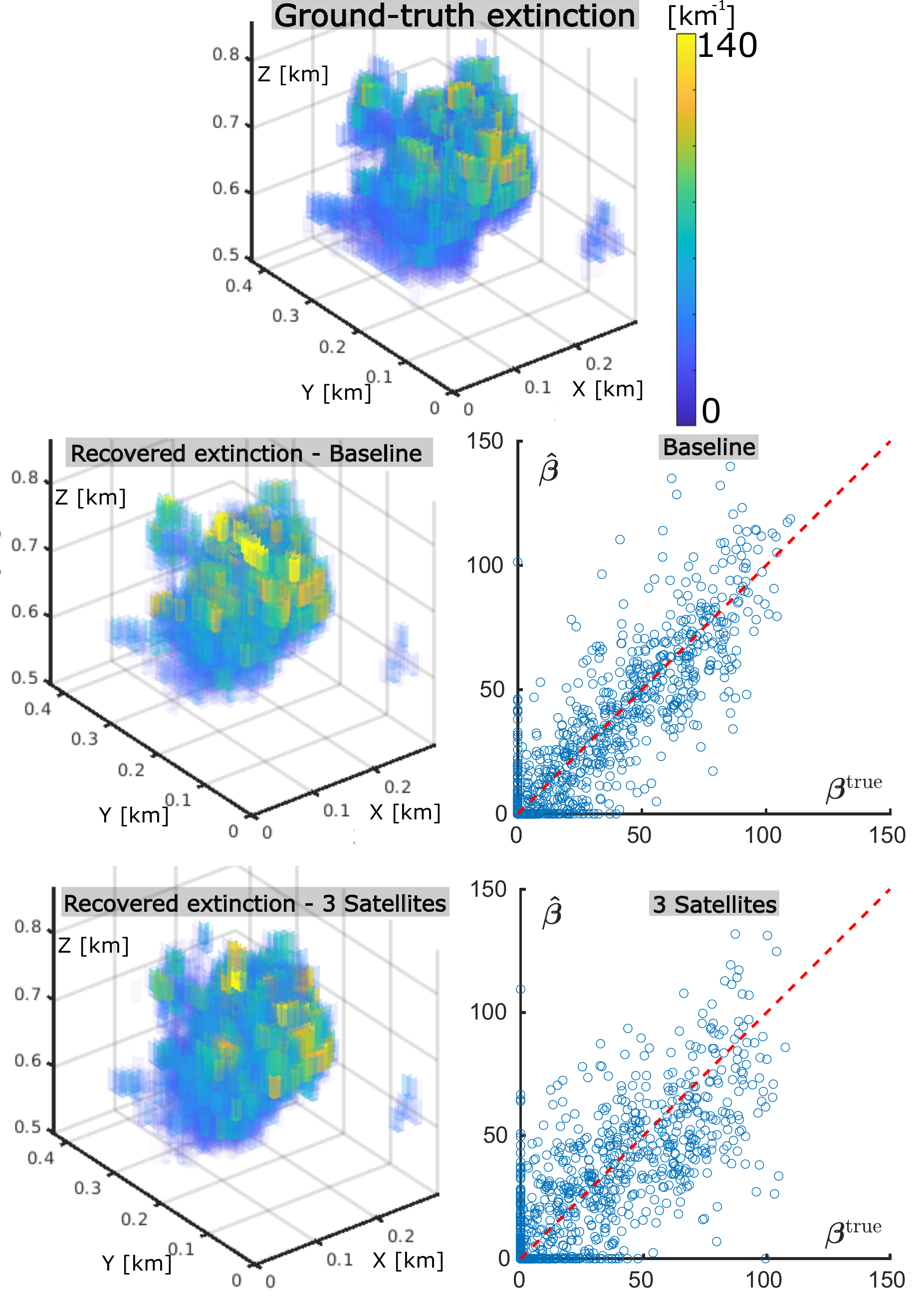}
  \vspace{-0.3cm}
  \caption{{\em Cloud~(i)}. Results of recovery by the {\tt Baseline} and {\tt Setup A} are compared to the ground-truth by a 3D presentation and scatter plots that use 20\% of the data points, randomly selected for display clarity.}
  \label{fig:sim_cloud1_3dvolumes}
\end{figure}

\begin{figure}
 \centering
  \includegraphics[width=1\linewidth]{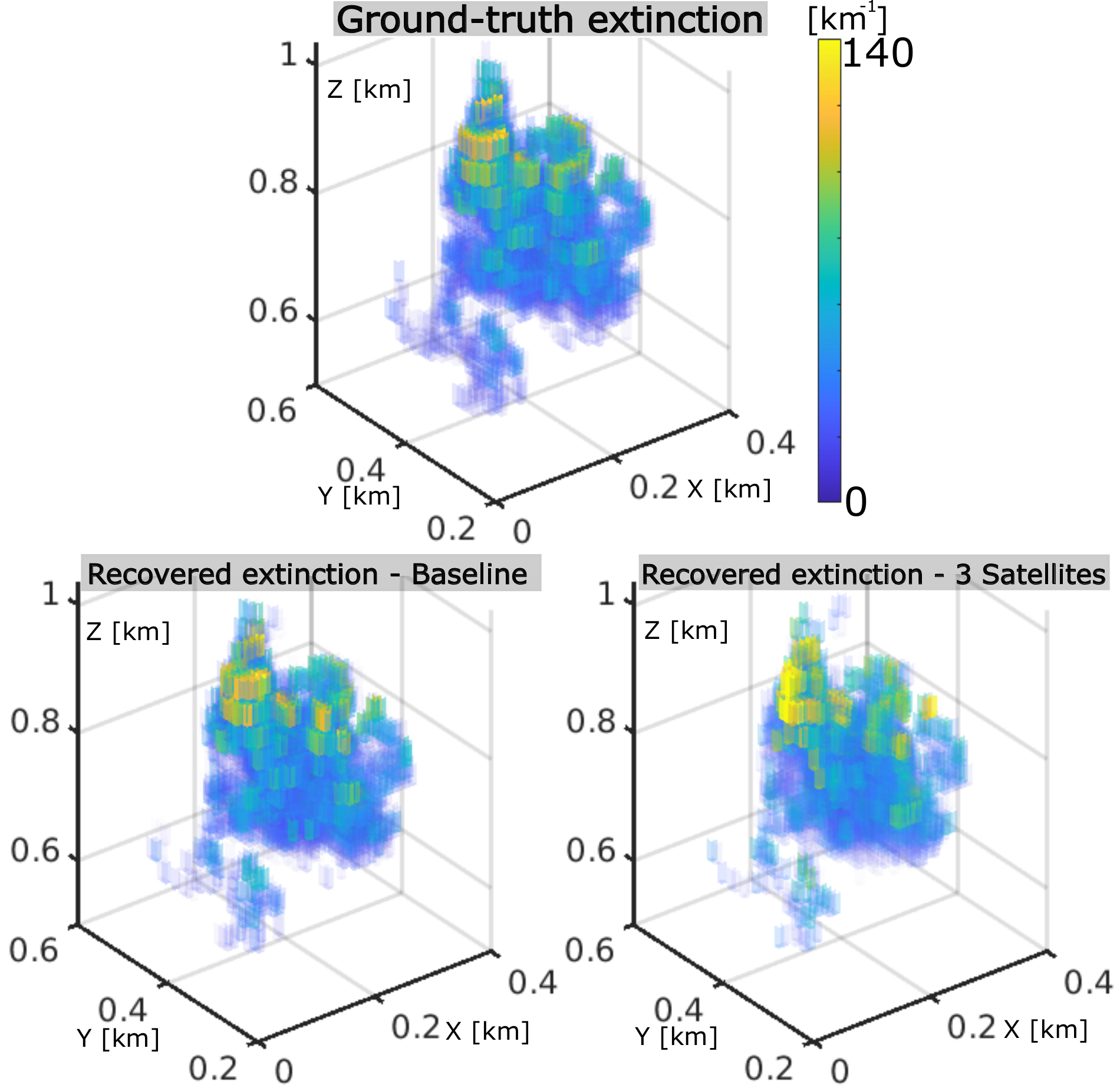}
  \vspace{-0.3cm}
  \caption{{\em Cloud~(ii)} Results of recovery by the {\tt Baseline} and {\tt Setup A} are compared to the ground-truth.}
  \label{fig:sim_cloud2_3dvolumes}
\end{figure}

We present results using two different time-varying clouds: {\em Cloud~(i)} has size $43\times 30 \times 45$ voxels (See Fig.~\ref{fig:sim_cloud1_3dvolumes}). {\em Cloud~(ii)} has size $60\times 40 \times 45$ voxels (See Fig.~\ref{fig:sim_cloud2_3dvolumes}). The voxel resolution is $10m\times 10m \times 10m$.

\subsection{Rendered Measurements}
\label{sec:MeasurementsRendering}
\begin{figure}
 \centering
  \includegraphics[width=1\linewidth]{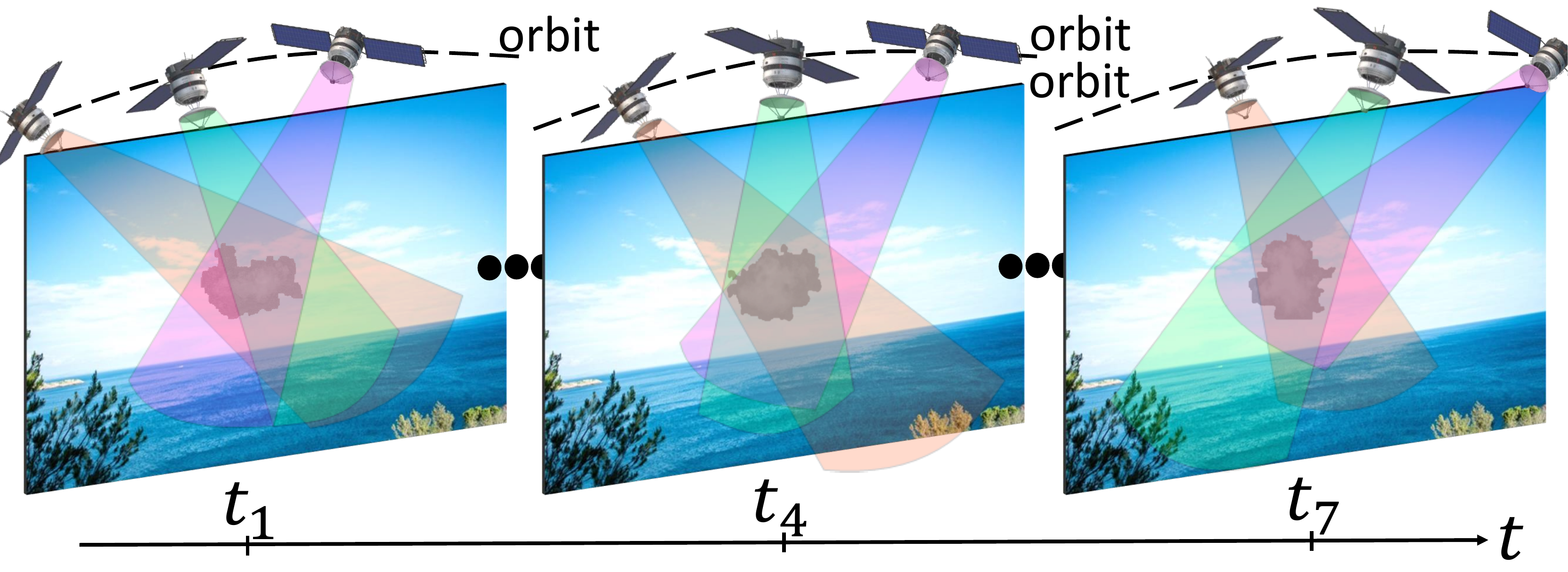}
  \vspace{-0.3cm}
  \caption{Illustration of {\tt Setup A}.}
  \label{fig:sat_illustration}
\end{figure}
Using the time-varying size distribution of the cloud droplets, 
Mie theory~\cite{bohren2008absorption} provides the spatiotemporal extinction field 
${\cal B}= \{ {\boldsymbol \beta}_{t'}\}_{t'\in {\cal T}}$ and scattering phase function. The scene is irradiated by the sun, whose illumination angle changes in time, relative to the Earth's coordinates, while cameras overfly the evolving cloud. The solar trajectory in Earth coordinates corresponds to Feb/03/2013 at 13:54:30 - 14:01:00 local time, around 38N 123W. We tested several types of imaging setups:\\

\noindent {\tt Setup A}: Three satellites orbit at $500{\rm km}$ altitude, one after the other. Their velocity is $7.35{\rm km}/{\rm sec}$. The orbital arc-length between nearest-neighboring satellites is $500{\rm km}$. At mid-time of the simulation, 
$t=(t_1+t_{N^{\rm state}})/2$, the setup is symmetric around the nadir direction. Then, the setup spans an angular range of $114^\circ$. Each satellite carries a perspective camera. The camera resolution is such that at nadir view, a pixel corresponds to  $10{\rm m}$ at sea level. Images are taken every  $10{\rm sec}$, during $60{\rm sec}$, i.e. 
$N^{\rm state}=7$. This setup is illustrated in Fig.~\ref{fig:sat_illustration}.\\

\noindent {\tt Baseline}: The baseline uses all the accumulated 21 viewpoints of {\tt Setup A}. However, all viewpoints here have perspective cameras that {\em simultaneously} acquire the cloud. In other words, this baseline is not prone to errors that stem from temporal sampling. The baseline is used for recovery only at time $t=(t_1+t_{N^{\rm state}})/2$.\\

\noindent {\tt Setup B}: This setup is similar to {\tt Setup A}, but it uses only two satellites. Thus at mid-time of the simulation, the setup spans a $57^\circ$ angular range. \\

\begin{figure}[t]
 \centering
  \includegraphics[width=1\linewidth]{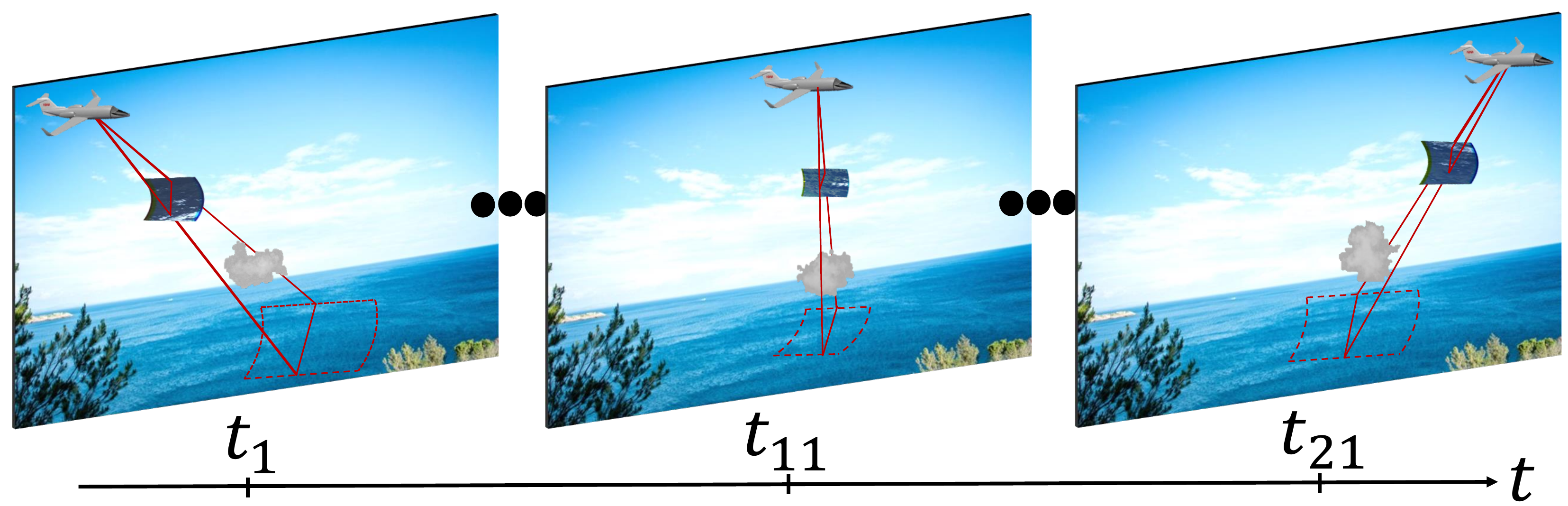}
  \vspace{-0.3cm}
  \caption{Illustration of {\tt Setup C}.  A domain is viewed at 21 pushbroom angles, sequentially.}
  \label{fig:single_platform_illustration}
\end{figure}

\noindent {\tt Setup C}:  A single camera, similar to the Multi-angle Spectro-Polarimeter Imager (AirMSPI)~\cite{diner2013airborne}, mounted on an aircraft flying $154^\circ$ relative to North at $20{\rm km}$ altitude.
Imaging has a pushbroom scan geometry, having $10{\rm m}$ spatial resolution at Nadir view and $\lambda=660{\rm nm}$ wavelength band.  AirMSPI scans view angles in a step-and-stare  mode~\cite{diner2013airborne}. 
Based on AirMSPI PODEX campaign~\cite{diner2013airbornePODEX}, we set 21 viewing angles along-track:  $\pm65^\circ,\ \pm62^\circ,\ \pm58^\circ,\ \pm54^\circ,\ \pm50^\circ,\ \pm44^\circ,\ \pm38^\circ,\ \pm30^\circ,\\ \pm21^\circ,\pm11^\circ$ off-nadir and 
$0^\circ$ (nadir). For example, three sample angles are illustrated in  Fig.~\ref{fig:single_platform_illustration}. 
It takes $\approx 1{\rm sec}$ to scan a cloud domain in any single view angle, during which the cloud and solar directions are assumed constant. Dynamics are noticeable {\em between} view angles.\\  

A spherical harmonic discrete ordinate method (SHDOM) code~\cite{evans2003improvements} provides the numerical forward model ${\cal F}$. Simulated measurements 
$\{ {\boldsymbol y}_{t} \}_{t\in {\cal T}}$ include noise. The noise model follows the AirMSPI sensors parameters~\cite{diner2013airborne,van2018calibration}. There, the sensor full-well depth is 200,000 photo-electrons, readout noise has a standard deviation of 20 electrons, and the overall readout is quantized to 9 bits. 

\subsection{Tomography Results} 
\label{sec:NumericalResults}

The rendered and noisy images served as input to 4D tomographic reconstruction. 
The voxel size in the recovery was set to $10 {\rm m}\times 10 {\rm m}$ horizontal, $25 {\rm m}$ vertical and $10 {\rm sec}$ resolution.
For parallelization, optimization ran on a computer cluster, where each computer core was dedicated to rendering a modeled image from a distinct angle. The optimization was initialized by 
$\{{\boldsymbol \beta}_t \}_{\tau \in {\cal T}} =1$km$^{-1}$.
Convergence was reached in several dozen iterations. Depending on the number of input images,  it took between minutes to a couple of hours to complete, in total.

\begin{figure}[t]
    \centering
     \includegraphics[width=0.8\linewidth]{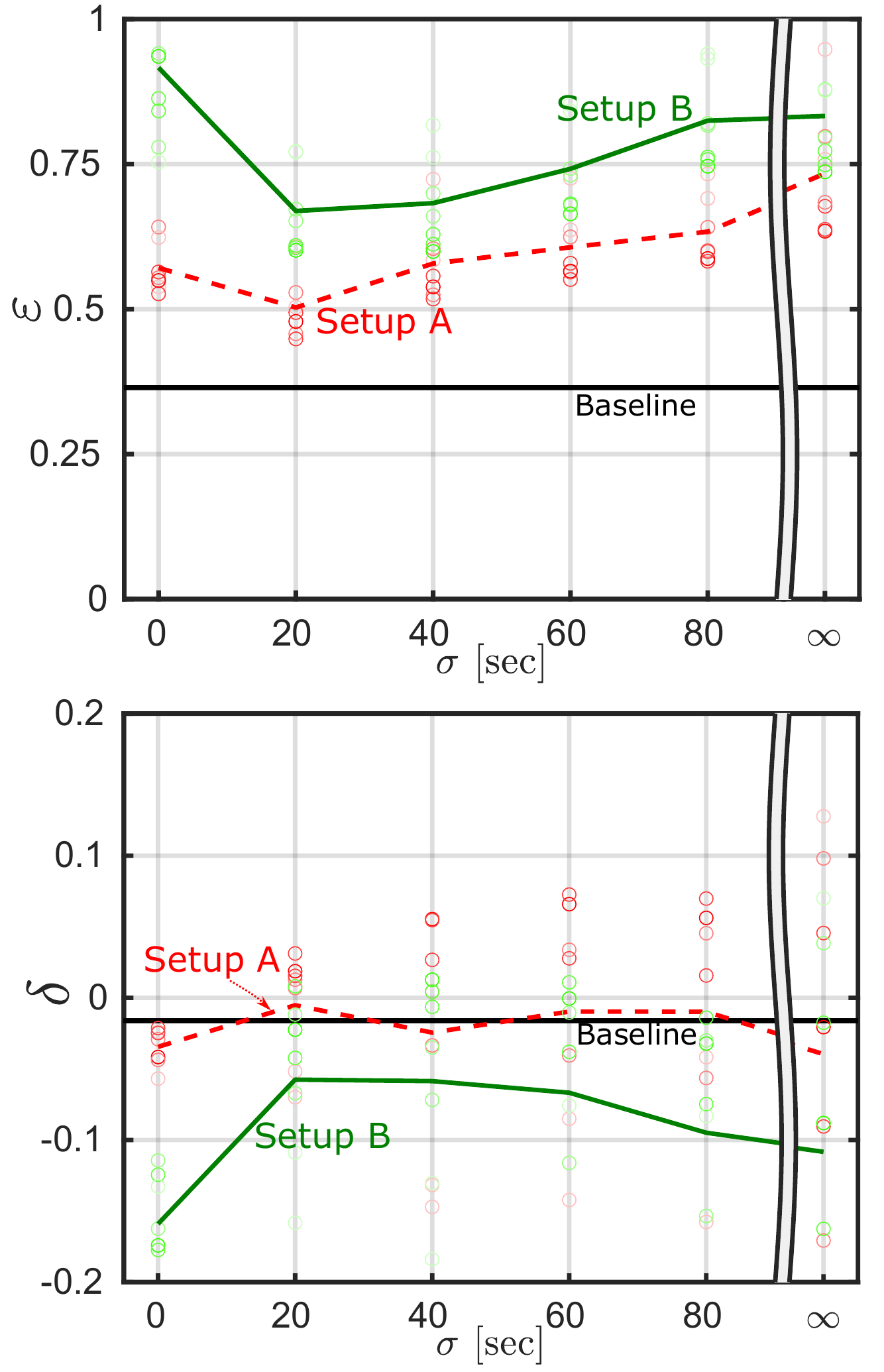}
      \caption{{\em Cloud~(i)}. The criteria of Eq.~(\ref{eq:MassError}) are marked by colored circles, whose saturation decays the farther the sampling time is from  $(t_1+t_{N^{\rm state}})/2$. The criteria in Eq.~(\ref{eq:epsilin}) marked by solid or dashed lines, with corresponding colors. The setting $\sigma=\infty$ refers to the solution by the state of the art, i.e. 3D static scattering tomography.} 
    \vspace{-0.3cm}
    \label{fig:error_sigma}
\end{figure}

From Sec.~\ref{sec:FrequencyAnalysis}, we assess that
a value $\sigma\sim 20{\rm sec}$ is natural. Indeed, this is supported numerically in the plots of $\varepsilon_t,\delta_t,\varepsilon,\delta$  for {\em Cloud~(i)} (Fig~\ref{fig:error_sigma}).
Analogous plots for {\em Cloud~(ii)} are presented in the Appendix. 
The 3D tomographic results using  {\tt Setup A} are shown in Figs.~\ref{fig:sim_cloud1_3dvolumes} and~\ref{fig:sim_cloud2_3dvolumes}, corresponding to  {\em Cloud~(i)} and {\em Cloud~(ii)}. 
Both illustrate the state at $t=(t_1+t_{N^{\rm state}})/2$. Recovery used $\sigma=20$sec. More results, particularly relating to {\tt Setup B} are shown in the Appendix. 

\begin{figure}[t]
      \centering \includegraphics[width=0.8\linewidth]{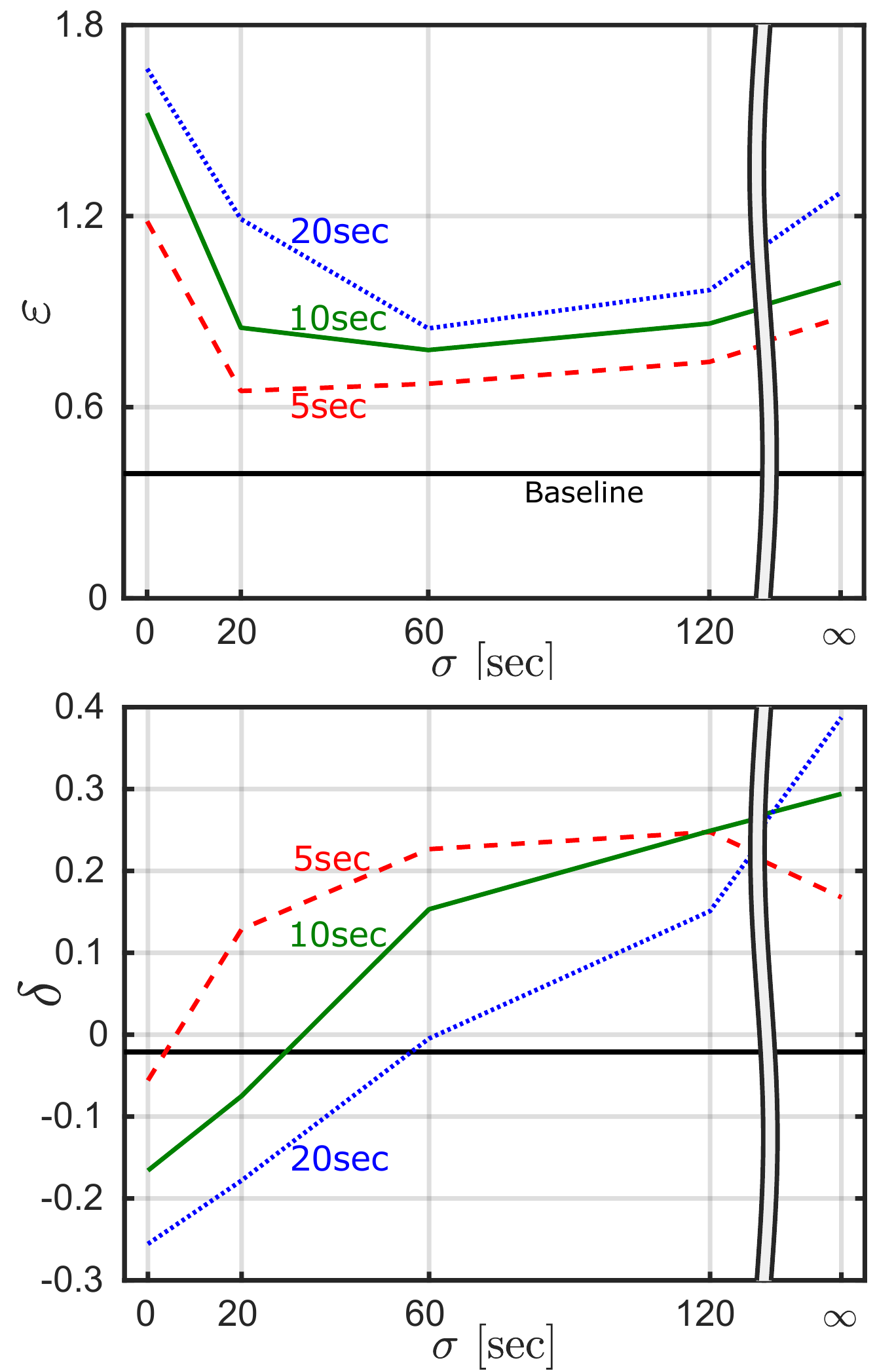} 
      \vspace{-0.3cm}
    \caption{{\tt Setup C}. Error measures~(\ref{eq:MassError}) of {\em Cloud~(i)}
    at time $t=(t_1+t_{N^{\rm state}})/2$, for different
    acquisition inter-angular temporal intervals.  
     The setting $\sigma=\infty$ refers to the solution by the state of the art, i.e. 3D static scattering tomography.}
    \label{fig:single_platform_errors}
\end{figure}

{\tt Setup C} uses a single platform, which is challenging.
Results depend significantly on how fast the aircraft flies, i.e. how long it takes to capture the cloud from a variety of angles (up to 21 angles).  Fig.~\ref{fig:single_platform_errors} compares the results for the recovery at $t=(t_1+t_{N^{\rm state}})/2$ for 
inter-angle time interval of 5sec, 10sec and 20sec. As expected, quality ($\varepsilon$) improves with velocity. Moreover, if the camera moves slowly (long time interval between angular samples), results improve by using a longer temporal support, observing the cloud from a wider angular range, despite its dynamics.
  
\section{Experiment: Real World AirMSPI Data} 
\label{sec:ResltdAirMSPI}

We follow the experimental approach of~\cite{levis2015airborne}, and use real-world data acquired by JPL's AirMSPI, which flies on board NASA’s ER-2.  The geometry is exactly as described in {\tt Setup C} in Sec.~\ref{sec:MeasurementsRendering}, including location and time. We examine an atmospheric domain of size $1.5{\rm km}\times 2{\rm km}\times 2{\rm km}$ in the East-North-Up coordinates. We discretized the domain to $80\times 80\times 80$ voxels. Because $N^{\rm states}=21$, the total number of unknowns is 10,752,000. 

The inter-angle time interval in this experiment is around 20sec. Based on Fig.~\ref{fig:single_platform_errors}, we set $\sigma = 60{\rm sec}$ here. We want to focus on dynamic tomography of the evolving cloud, and not on global motion due to wind in the cloud field. Hence, we used the pre-processing approach of~\cite{levis2015airborne} to align the cloud images. Additionally, the ground albedo is estimated to be 0.04.  The pre-processing and albedo estimation are described in the Appendix. 

\begin{figure}[t]
 \centering
  \includegraphics[width=1\linewidth]{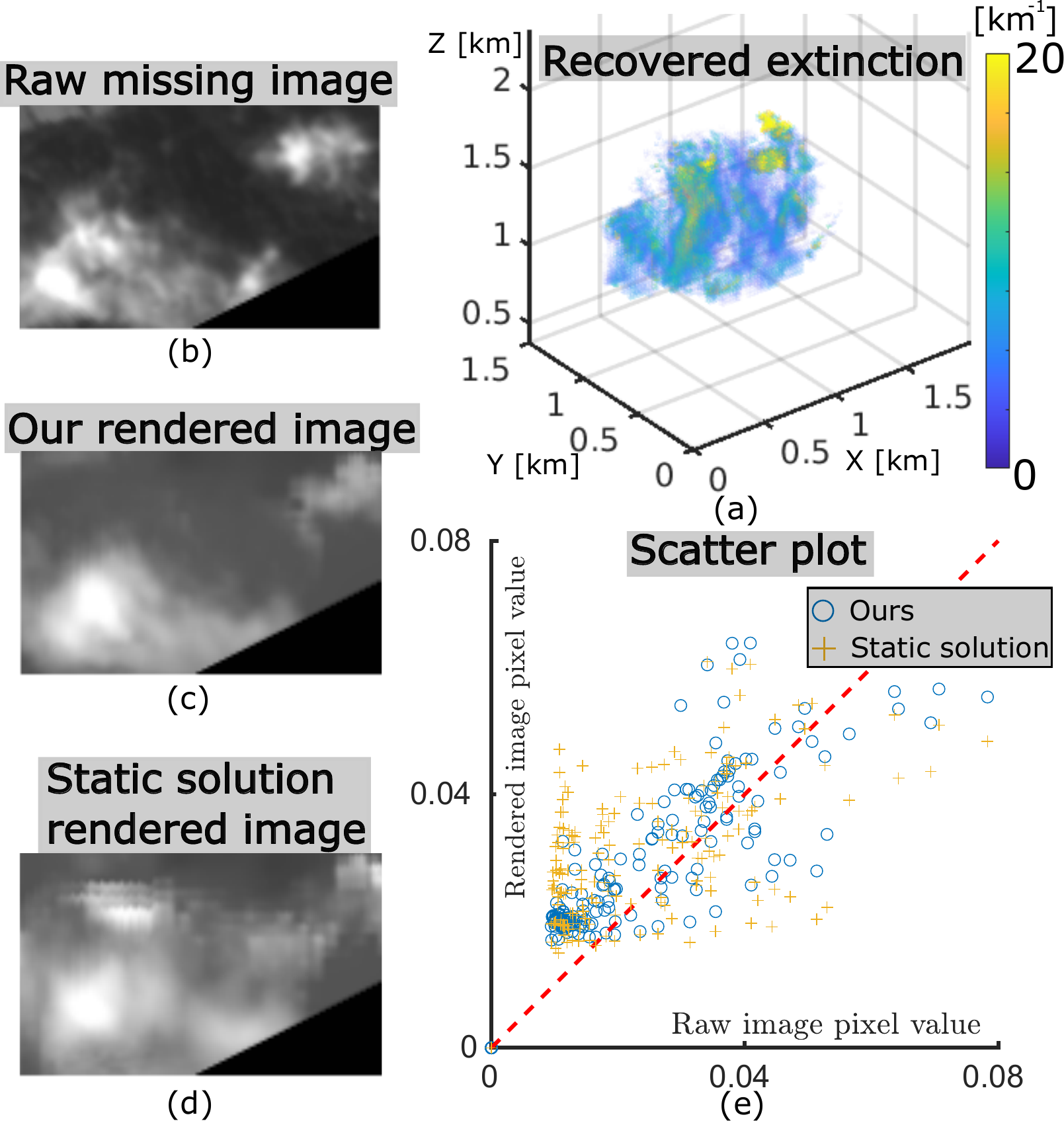}
  \vspace{-0.3cm}
  \caption{Recovered 3D extinction field using real data (a). A raw AirMSPI nadir image (b). Corresponding rendered views of a cloud, that was estimated using data that had excluded the nadir, either by our 4D CT approach (c) or current static 3D CT (d). Gamma correction was applied on (b,c,d) for display clarity. (e) A scatter plot of  rendered vs. raw AirMSPI images at nadir.}
  \label{fig:airmspi_results}
\end{figure}

A recovered volumetric reconstruction for one time instant is displayed in Fig.~\ref{fig:airmspi_results}. We have no ground-truth for the cloud content in this case. Hence we check for consistency using cross-validation. For this, we excluded the nadir image (Fig.~\ref{fig:airmspi_results}b) from the recovery process. Thus tomography used 20 out of the 21 raw views. Afterward, we placed the recovered cloud in SHDOM physics-based rendering~\cite{evans2003improvements}, to generate  
the missing nadir view. The result is then compared to the ground-truth missing view. Fig.~\ref{fig:airmspi_results} compares the result of this process for two solutions: our 4D tomographic solution, and the state-of-the-art, i.e., 3D static scattering tomography. 

The same cross-validation process was repeated for the $\pm54^\circ$ view angles. Quantitatively, we measure the fitting error using Eq.~(\ref{eq:Et}). The results are summarized in Table~(\ref{tab:airmspi}).
\begin{table}[t]
\begin{center}
\begin{tabular}{|l|c|c|c|}
\hline
  &$+54^\circ$ view &  nadir view &  $-54^\circ$ view \\
\hline\hline
Ours & 0.96& 0.38& 0.24 \\
Static solution & 1.73& 0.94 & 0.61\\
\hline
\end{tabular}
\end{center}
      \vspace{-0.3cm}
    \caption{Analysis of empirical data in different view angles. Quantitative fit (\ref{eq:Et}) of our 4D result to the data, as compared to the error of state-of-the-art static 3D CT.}
\label{tab:airmspi}
\end{table}


\section{Discussion}
\label{sec:discuss}

We derive a framework for 4D CT of dynamic objects that scatter, using moving cameras. The natural temporal evolution of an object indicates the temporal and angular sampling  needed for a good reconstruction.  Given these conditions, 4D CT recovery can be done, even with a small number of cameras. We believe that our approach can be relevant in additional tomographic setups~\cite{gumbel2020mats} that rely on radiative transfer.
Some elements of this work are generic, beyond scattering CT. Thus,  it is worth applying the approach to other tomographic  modalities. Our findings can significantly improve various research fields, including bio-medical CT, flow imaging and atmospheric sciences.

\section*{Acknowledgment}
We are grateful to Aviad Levis and Jesse Loveridge for the  pySHDOM code and for being responsive to questions about it. We are grateful to Vadim Holodovsky and Omer Shubi for helping in setting elements of the code. We thank the following people: Ilan Koren, Orit Altaratz  and Yael Sde-Chen for useful discussions and good advice; Danny Rosenfeld for pointing out Ref.~\cite{fujita1986mesoscale} to us, Johanan Erez, Ina Talmon and Daniel Yagodin for technical support. Yoav Schechner is the Mark and Diane Seiden Chair in Science at the Technion. He is a Landau Fellow - supported by the Taub Foundation.
His work was conducted in the Ollendorff Minerva Center. Minvera is funded
through the BMBF. This project has received funding from the European Union’s Horizon 2020 research and innovation programme under grant agreement No 810370-ERC-CloudCT.

\appendix
\section*{Appendices}
    We now present several appendices.
    Appendix~\ref{sec:preprocessing} elaborates on  pre-processing which is applied to real world measurements that are presented in Sec.~\ref{sec:ResltdAirMSPI}. This data was collected by the AirMSPI instrument.  Appendix~\ref{sec:simulations} provides an additional example of the bandwidth of the power spectrum of a cloud and more simulation results. Appendix~\ref{sec:complexity} analyzes the computational complexity of the method. 


\section{Pre-processing Real World Data}
\label{sec:preprocessing}

\begin{figure}[t]
    \centering
     \includegraphics[width=1\linewidth]{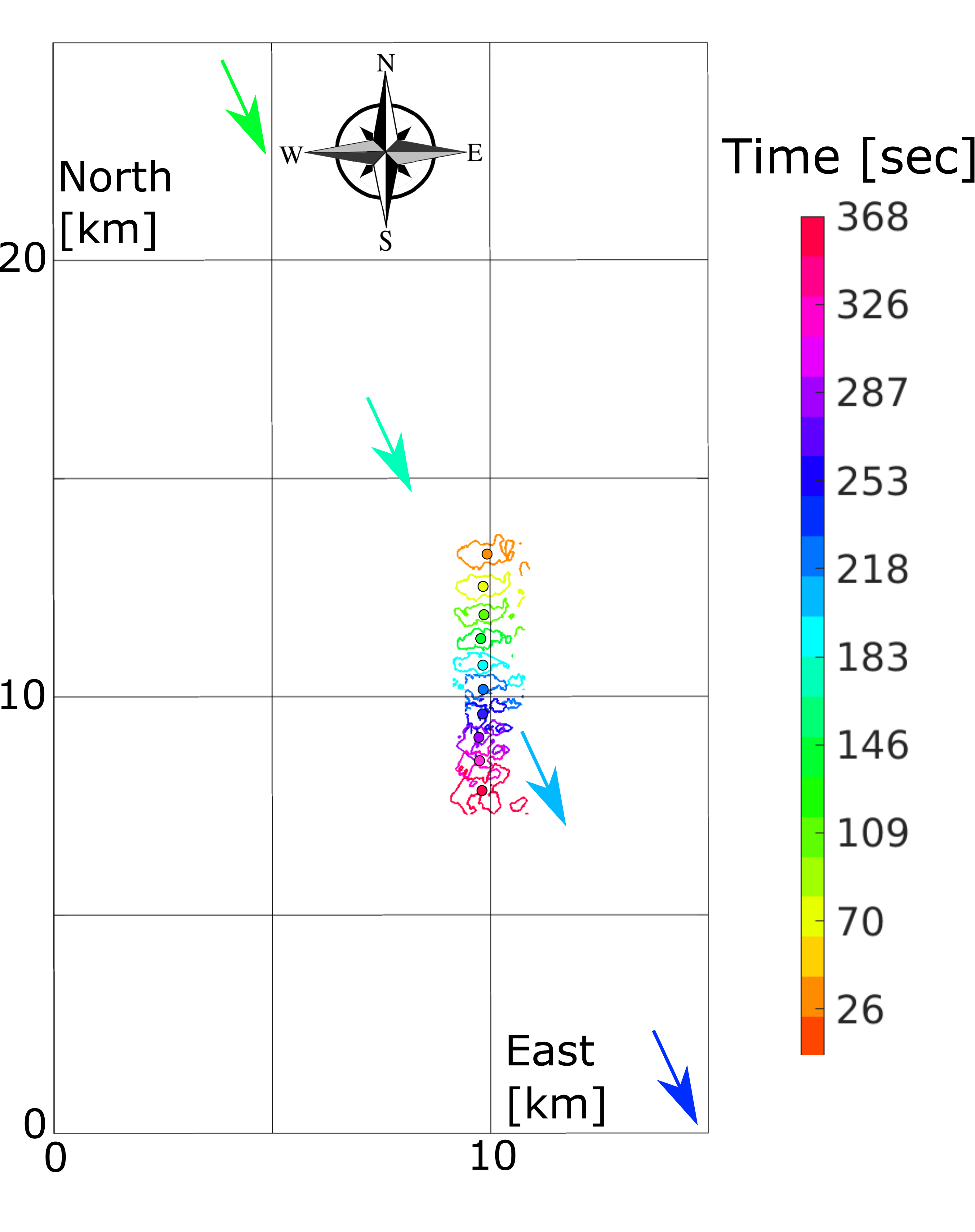}
      \caption{{\small Geometry of the AirMSPI real world setup which led to the data  presented in Sec.~\ref{sec:ResltdAirMSPI}. The color represents the locations of the cloud and the AirMSPI instrument in the different time states. The cloud's outer contour and its corresponding center of mass, marked in a circle, are presented per state. The AirMSPI location and velocity are marked by arrows.  The arrows point to the AirMSPI  flight direction azimuth of $154^\circ$ relative to the North. Due to the domain size, not all AirMSPI locations are illustrated here. Due to wind, the cloud moves at 57km/h in azimuth $182^\circ$ relative to the North.  } }
    \label{fig:airmspi_geometry}
\end{figure}

Sec.~\ref{sec:ResltdAirMSPI} presents results using real world measurements. The data were acquired by the AirMSPI instrument. As explained in Sec.~\ref{sec:ResltdAirMSPI}, while AirMSPI flies, clouds move due to wide-scale wind at their altitude. The geometry of AirMSPI's path and the cloud drift during the experiment is presented in Fig.~\ref{fig:airmspi_geometry}. 
In order to eliminate the influence of wide-scale wind,  a registration process of the cloud images is done. Moreover, for tomographic recovery, we need to have an assessment of the Earth surface albedo, under the clouds. This section describes how pre-processing estimates the wind and albedo.  

\subsection{Wind Estimation}
\label{sec:WindEstimation}

\begin{figure}[t]
    \centering
     \includegraphics[width=1\linewidth]{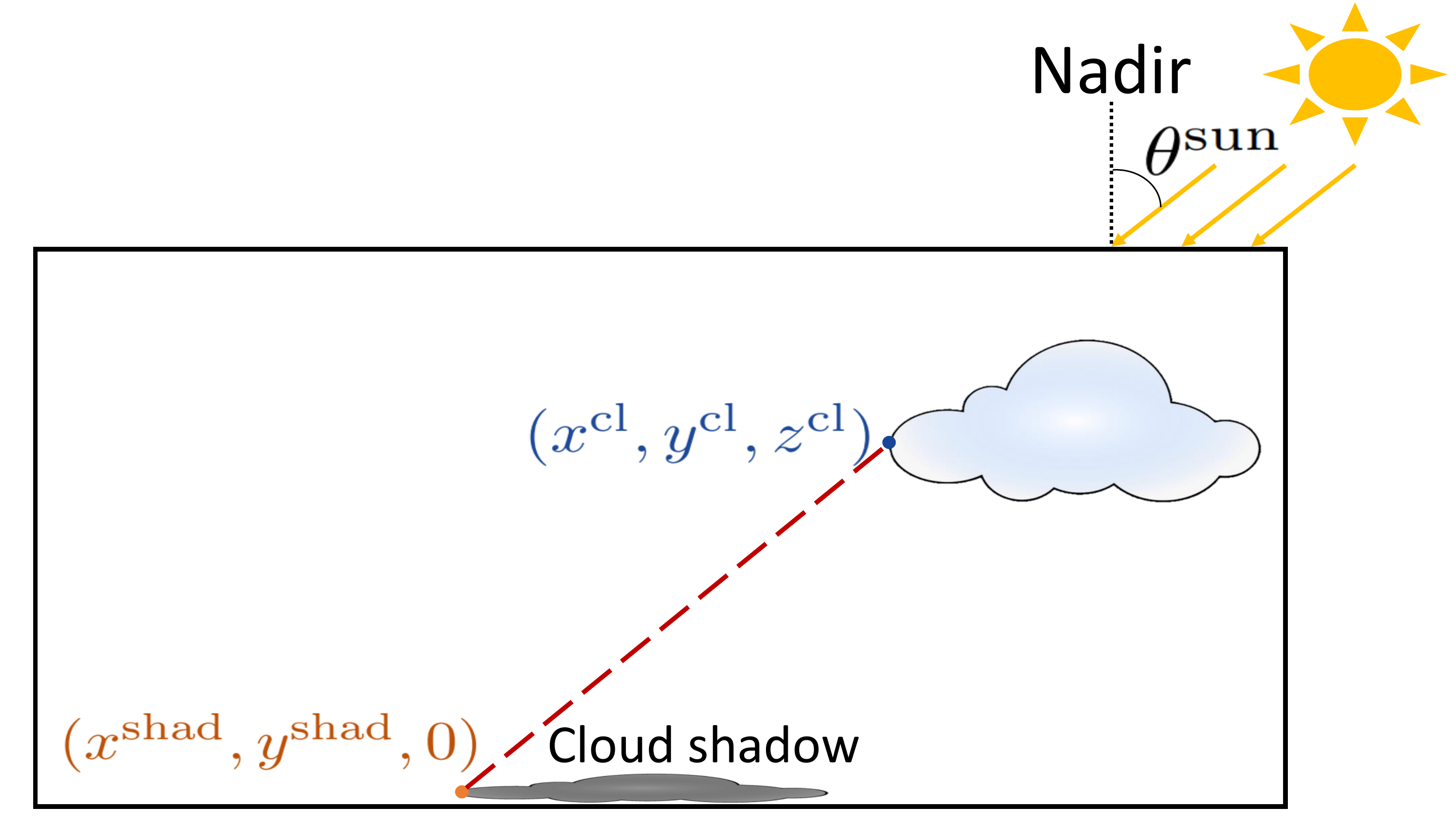}
      \caption{Illustration of estimation of cloud altitude using its shadow.} 
    \label{fig:cloud_height}
\end{figure}

Clouds are segmented from the surface automatically~\cite{velasco1979thresholding}. Cloudy pixels are then used to estimate the cloud center of mass in each image~\cite{levis2015airborne}. A registration of these centers of mass can be done by triangulation.
However, triangulation of images of a moving object using a translating camera  has an inherent ambiguity. This ambiguity can be solved if the cloud height is known. In this work, we assess this altitude of a cloud by its shadow~\cite{abrams2013episolar,liasis2016satellite,hatzitheodorou1988optimal}. Let $(x^{\rm cl}, y^{\rm cl}, z^{\rm cl})$ and $(x^{\rm shad}, y^{\rm shad}, 0)$ be a point in a cloud and its corresponding shadow point on the earth surface, respectively (see Fig.~\ref{fig:cloud_height}). 
Let $r^{\rm shad}=\sqrt{(x^{\rm cl} - x^{\rm shad})^2+(y^{\rm cl} - y^{\rm shad})^2}$. We obtain $x^{\rm cl}, y^{\rm cl}, x^{\rm shad}$ and  $y^{\rm shad}$ from the AirMSPI images.  Given the solar zenith angle relative to the nadir $\theta^{\rm sun}$, the altitude  $z^{\rm cl}$ satisfies
\begin{equation}
   z^{\rm cl} = \frac{r^{\rm shad}}{\tan(\theta^{\rm sun})}
   \;.
   \label{eq:cloud_height}
\end{equation}
For the example shown in Sec.~\ref{sec:ResltdAirMSPI}, we estimated the cloud base height as $\approx$500m and its top at $\approx$1100m. Indeed taking MODIS/AQUA~\cite{NASAsite} retrievals of cloud top heights, indicates that the clouds' top in the region\footnote{This data applies over the coast of California, 38N 122W,on  Feb/03/2013 at 13:30 local time.}  does not exceed 1000m, which makes our approximation reasonable.

We approximate the cloud horizontal velocity by projecting the images from the locations of the camera  to the altitude of $z^{\rm cl}$. From the center of mass of these images, we assess the velocity. We register the camera locations so the projections of the center of mass of all images intersect at the same point at altitude of $z^{\rm cl}$. The images and the registered camera locations are the input for 4D recovery.

\subsection{Surface Albedo Estimation}
\label{sec:albedoest}

3D radiative transfer calculations require the surface albedo. We use non-cloudy pixels to estimate the albedo. Let ${\cal Y}$ be a set of non-cloudy pixels.  We estimate the surface albedo $a$ as,
\begin{equation}
   {\hat a} = \arg\! \min_a \sum_{y\in{\cal Y}} ||y-{\cal F}(\beta^{\rm air};a)||_2^2
   \;.
   \label{eq:albedo}
\end{equation}
Here ${\cal F}(\beta^{\rm air};a)$ is a rendering (forward) model where the surface albedo is set to be $a$ and the atmospheric medium contains no clouds. That is, sunlight interacts only with the air and the surface. Scattering by air is assumed to be known~\cite{gordon1988exact,wang2005refinement}. The optimization problem is solved by the Brent minimization method~\cite{brent2013algorithms}, implemented by the SciPy package~\cite{scipy}. For the example shown in Sec.~\ref{sec:ResltdAirMSPI}, the surface albedo is estimated to be 0.04.


\section{Additional Simulations}
\label{sec:simulations}
\subsection{Cloud Temporal Spectrum}
\label{sec:spectrum}

Sec.~\ref{sec:FrequencyAnalysis} indicates that the temporal power spectrum of a convective cloud at 10m resolution is effectively limited. Thus, a temporal sampling period of 25[sec] or shorter is required. We assess this in an additional cloud simulation. We conducted a single cloud simulation in high resolution, using small changes, relative to the simulation  described in Sec.~\ref{sec:CloudFieldSimulation}. The simulation parameters and setting are similar. However, the perturbation that initiates the convection and turbulent flow  has a  smaller horizontal size. This creates a smaller cloud with a horizontal width of $\approx$400m. This cloud is more  sensitive to mixing and evaporation than the cloud in Sec.~\ref{sec:CloudFieldSimulation} whose width is $\approx$800m. Because mixing with the environment is more intense here, the clouds' growth is inhibited. It can not  exceed a height of 1400m, compared to a 2000m ceiling of the cloud in Sec.~\ref{sec:CloudFieldSimulation}.

\begin{figure}[t]
    \centering
     \includegraphics[width=1\linewidth]{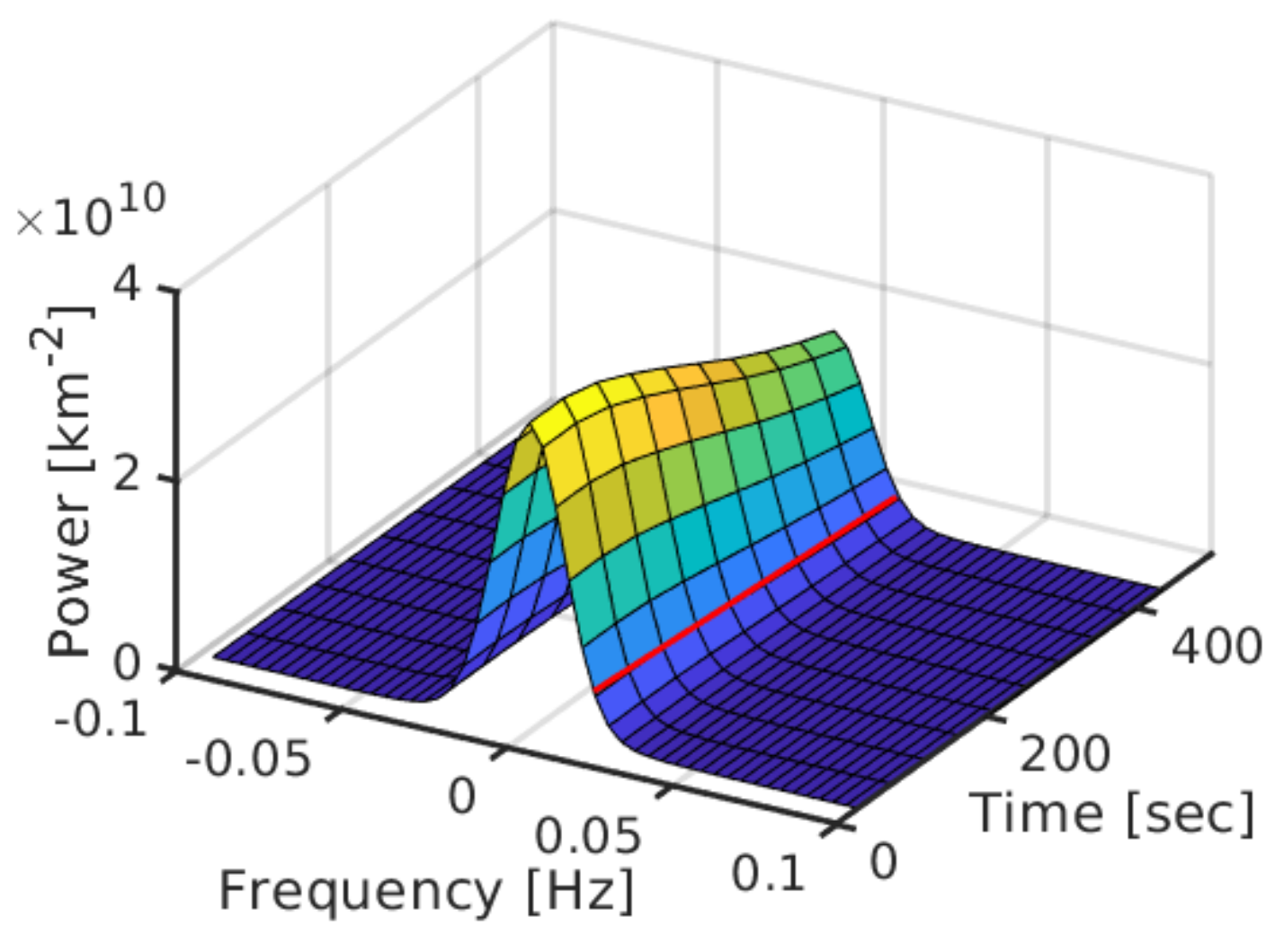}
      \caption{A cutoff frequency $\approx 1/70 {\rm [Hz]}$, within which 95\% of the signal power is contained, is marked in red.} 
    \label{fig:stft_sup}
\end{figure}

Using the same process described in Sec.~\ref{sec:FrequencyAnalysis}, the temporal power spectrum of the cloud is presented in Fig.~\ref{fig:stft_sup}. 
The cutoff frequency is $\approx1/70$[Hz], and it is not sensitive to the evolving stages of the cloud. Here the required temporal sampling period is 35[sec] or shorter. This is more tolerable compared to the temporal sampling period in Sec.~\ref{sec:FrequencyAnalysis}.

\subsection{Additional Tomography Results}
\label{sec:results}

\begin{figure}[th!]
    \centering
     \includegraphics[width=0.85\linewidth]{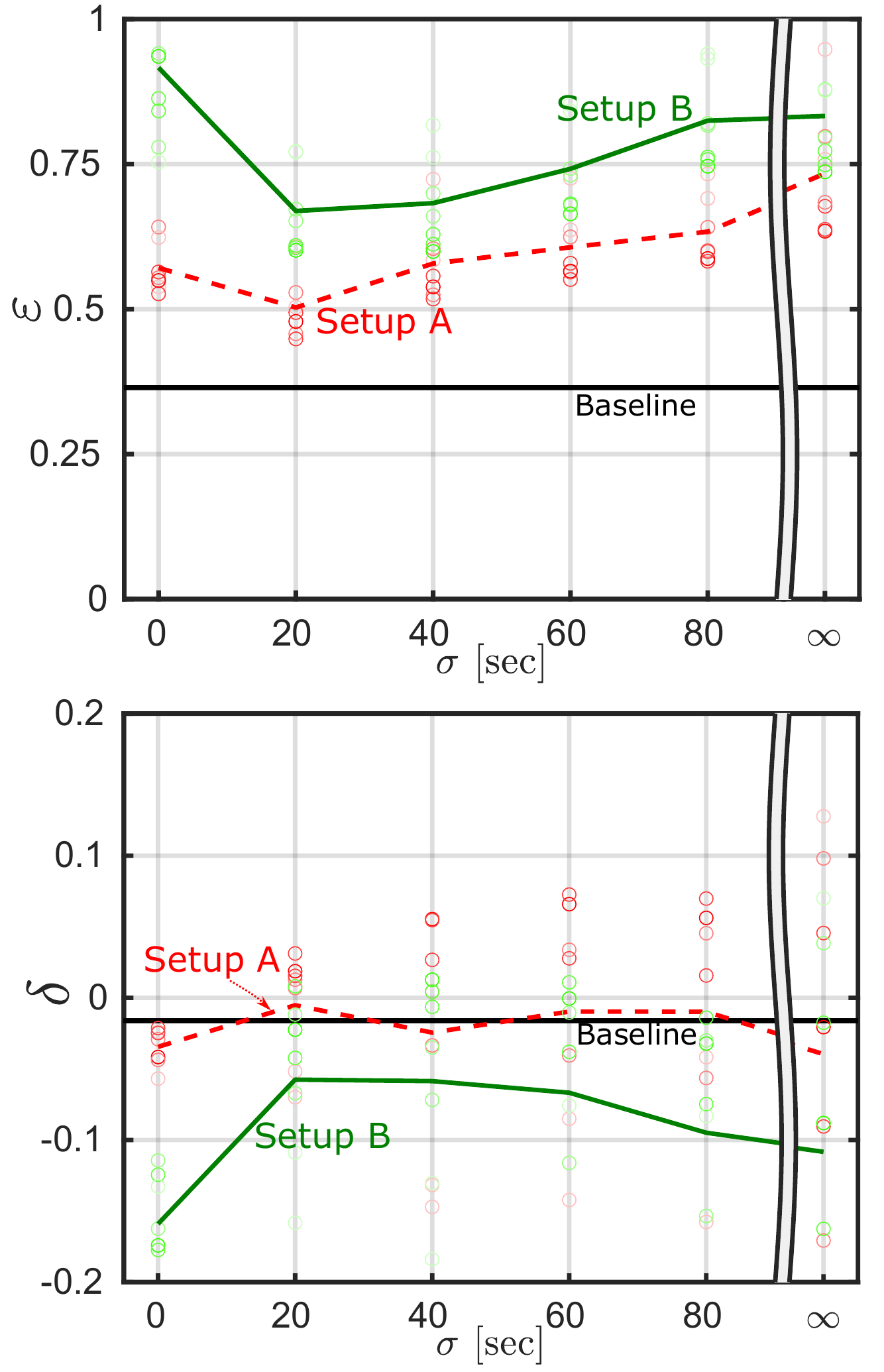}
      \caption{{\small {\em Cloud~(ii)}. The criteria of Eq.~(\ref{eq:MassError}) are marked by colored circles, whose saturation decays the farther the sampling time is from  $(t_1+t_{N^{\rm state}})/2$. The criteria in Eq.~(\ref{eq:epsilin}) are marked by solid or dashed lines, with corresponding colors. The setting $\sigma=\infty$ refers to the solution by the state of the art, i.e. 3D static scattering tomography.}} 
    \label{fig:error_sigma_sup}
\end{figure}

Recall that our method is demonstrated on two simulated clouds, {\em Cloud~(i)} and {\em Cloud~(ii)}, using several types of imaging setups: {\tt Setup A}, {\tt Setup B} and {\tt Baseline}.  Moreover, recall the error criteria as Eqs.~(\ref{eq:MassError},\ref{eq:epsilin}).
Fig.~\ref{fig:error_sigma_sup} shows $\varepsilon_t,\delta_t,\varepsilon,\delta$  for {\em Cloud~(ii)}. 
It reinforces the assessment that a value $\sigma\sim 20{\rm sec}$ is natural, as explained in Sec.~\ref{sec:FrequencyAnalysis}.
\begin{figure*}[t]
		\def\svgwidth{1\linewidth}
	\centering
	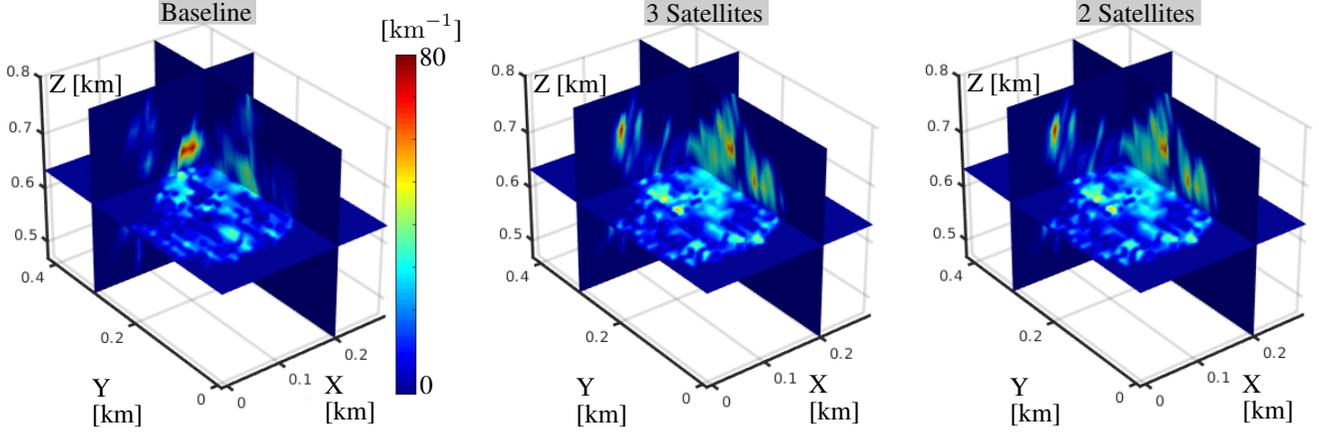
  \caption{{\em Cloud~(i)}. 3D cut-sections of the error $|{ \beta}^{\rm true}_t({\bf x}) - {\hat { \beta}}_t({\bf x})|$ at $t=(t_1+t_{N^{\rm state}})/2$  for {\tt Baseline}, {\tt Setup A} and {\tt Setup B}.}
    \vspace{0.3cm}
  \label{fig:cloud1_sctions}
\end{figure*}
\begin{figure*}[t]
		\def\svgwidth{1\linewidth}
	\centering
	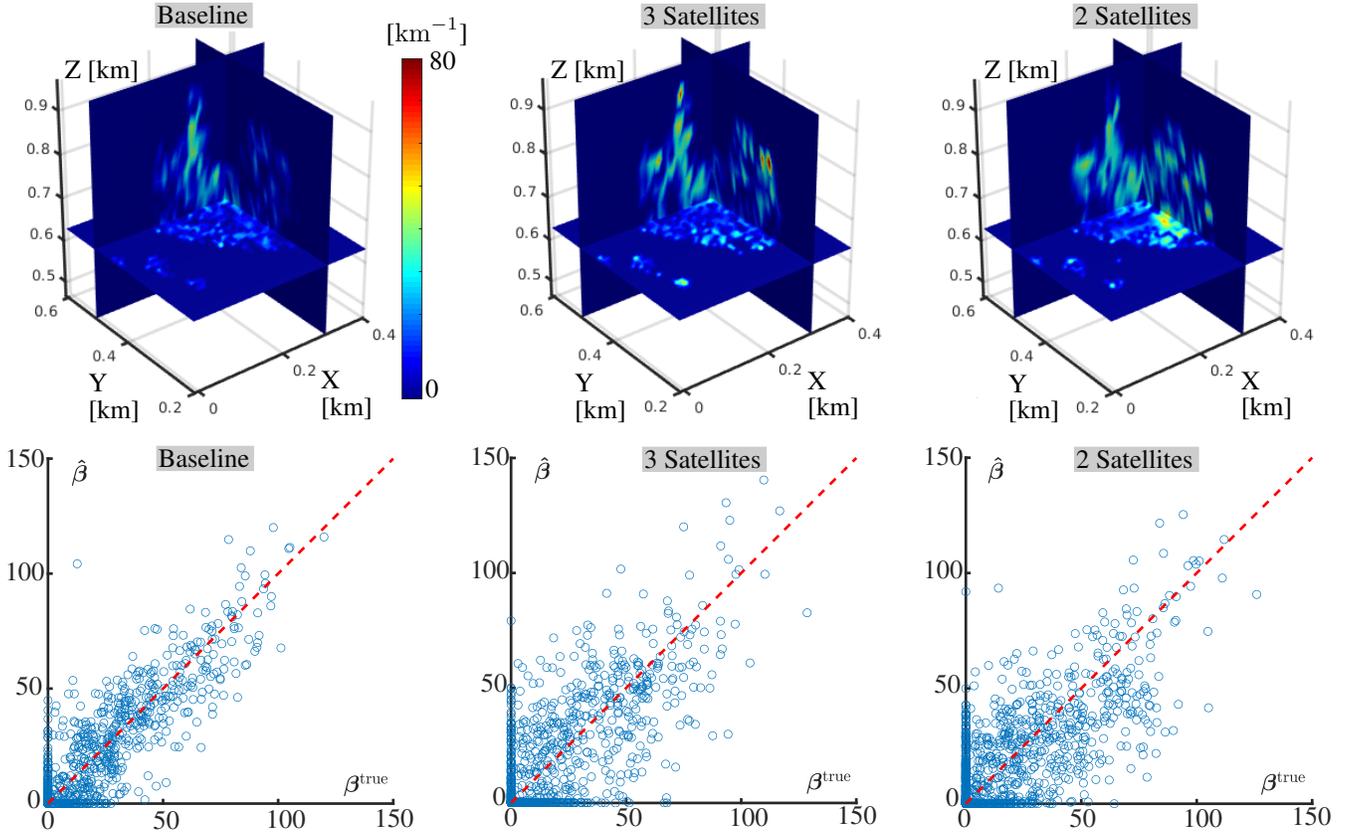
  \caption{{\em Cloud~(ii)}  comparison for {\tt Baseline}, {\tt Setup A} and {\tt Setup B}. [Top] 3D cut-sections of the error $|{ \beta}^{\rm true}_t({\bf x}) - {\hat { \beta}}_t({\bf x})|$ at $t=(t_1+t_{N^{\rm state}})/2$. [Bottom] Scatter  plots  that  use randomly selected  20\%  of  the  data points,  for display clarity.}
  \vspace{0.3cm}
  \label{fig:cloud2_sctions}
\end{figure*}
Figs.~\ref{fig:cloud1_sctions}  and~\ref{fig:cloud2_sctions} respectively visualize the results of   {\em Cloud~(i)} and  {\em Cloud~(ii)}.  The 3D cut-sections of the error $|{\beta}^{\rm true}_t({\bf x}) - {\hat { \beta}}_t({\bf x})|$ at $t=(t_1+t_{N^{\rm state}})/2$  is presented for {\tt Setup A}, {\tt Setup B} and {\tt Baseline} in Figs.~\ref{fig:cloud1_sctions} and~\ref{fig:cloud2_sctions}[Top]. Fig.~\ref{fig:cloud2_sctions}[Bottom] uses scatter plots to compare the ground-truth to the results obtained by either the {\tt Baseline}, {\tt Setup A} or {\tt Setup B}.

\section{Computational Complexity}
\label{sec:complexity}
The time complexity for solving the 4D CT inverse problem, (Eq.~\ref{eq:dynamic_inverse}) is governed by the gradient calculation.
Recall the formulation of the approximate gradient,
\begin{equation}
      {\bf g}_t({\cal B}) = \sum_{t'\in {\cal T}}
      w_{t'}(t)
     \left[ 
        {\cal F} \left({\boldsymbol \beta}_{t'} \right)  - {\boldsymbol y}_{t'} 
     \right]      
     \frac{\partial 
          {\cal F} \left({\boldsymbol \beta}_{t'} \right)
          }
          {\partial {\boldsymbol \beta}_{t'}}
       \;.  
   \label{eq:pgrad_sup}
\end{equation}
Computing the Jacobian 
$     \partial {\cal F} \left({\boldsymbol \beta}_{t'} \right)
        /
         \partial {\boldsymbol \beta}_{t'}$
is complex, thus it is established numerically by a surrogate function  that  evolves  through  iterations~\cite{levis2015airborne,loeub2020monotonicity}. Calculating the gradient includes two dominant time-consuming processes that are executed in alternation. The first process calculates the forward model for the $N^{\rm state}$ cloud states $\left\{ {\cal F} \left( {{\boldsymbol \beta}}_{t'} \right) \right\}_{t' \in {\cal T}}$. The second process sums over the entire set of measurements, which does not depend on the number of cloud states that we seek to recover.

A spherical harmonic discrete ordinate method (SHDOM) code is used for computing the numerical forward model ${\cal F}(\cdot)$ and the Jacobian. SHDOM iteratively updates the estimation of 3D radiation fields until convergence. Calculating the forward model for the $N^{\rm state}$ cloud states can be done in parallel. Thus, the time complexity is governed by the temporal state for which  the SHDOM code takes  the longest time to compute the forward model. By calculating the forward model for all cloud states in parallel, the time complexity of gradient calculation is insensitive to the number of cloud states $N^{\rm state}$ . 

As a numerical example, we used 20 iterations of the L-BFGS-B optimization. Using measurements of {\em Cloud~(i)} acquired by {\tt Setup A},  the run-time of the solution  by our method was 501[sec]. The static solution took 301[sec]. In both, the computer was Intel\textregistered \ Xeon\textregistered \ Gold 6240 CPU @ 2.60GHz with 72 cores. Although our method recovers $N^{\rm state}=7$ times more voxels, the run-time is less than twice that of the static solution. The time difference is caused by overheads of saving and loading larger data with our method, and nonoptimal task division for the cores.

{\small
\bibliographystyle{ieee_fullname}
\bibliography{egbib}
}


\end{document}